\documentclass[acmtog]{acmart}

\usepackage{booktabs} 

\usepackage[ruled]{algorithm2e} 
\usepackage{multirow}
\usepackage{subcaption}
\usepackage{colortbl}
\usepackage{xcolor}

\SetAlFnt{\small}
\SetAlCapFnt{\small}
\SetAlCapNameFnt{\small}
\SetAlCapHSkip{0pt}

\newtoggle{arXiv}
\toggletrue{arXiv}

\acmJournal{TOG}

\definecolor{rred}{RGB}{245, 152, 153}
\definecolor{oorange}{RGB}{253, 205, 154}
\definecolor{yyellow}{RGB}{248,244,140}

\makeatletter
\DeclareRobustCommand\onedot{\futurelet\@let@token\@onedot}
\def\@onedot{\ifx\@let@token.\else.\null\fi\xspace}
\def\eg{e.g\onedot} 
\def\ie{i.e\onedot}

\def\wrt{wrt\onedot}

\makeatother

\newcommand{\boldparagraph}[1]{\vspace{0.2cm}\noindent{\bf #1:} }

\definecolor{darkgreen}{rgb}{0,0.7,0}

\begin{document}
\title{Relightable Full-Body Gaussian Codec Avatars}

\author{Shaofei Wang}
\authornote{Work was done during an internship at Meta}
\affiliation{%
  \institution{ETH Zürich}
  \country{Switzerland}}
\email{shaofei.wang@inf.ethz.ch}
\author{Tomas Simon}
\affiliation{%
  \institution{Codec Avatars Lab, Meta}
  \country{USA}}
\email{tsimon@meta.com}
\author{Igor Santesteban}
\affiliation{%
  \institution{Codec Avatars Lab, Meta}
  \country{USA}}
\email{igor.santesteban@gmail.com}
\author{Timur Bagautdinov}
\affiliation{%
  \institution{Codec Avatars Lab, Meta}
  \country{USA}}
\email{timurb@meta.com}
\author{Junxuan Li}
\affiliation{%
  \institution{Codec Avatars Lab, Meta}
  \country{USA}}
\email{junxuanli@meta.com}
\author{Vasu Agrawal}
\affiliation{%
  \institution{Codec Avatars Lab, Meta}
  \country{USA}}
\email{vasuagrawal@meta.com}
\author{Fabian Prada}
\affiliation{%
  \institution{Codec Avatars Lab, Meta}
  \country{USA}}
\email{fabianprada@meta.com}
\author{Shoou-I Yu}
\affiliation{%
  \institution{Codec Avatars Lab, Meta}
  \country{USA}}
\email{shoou-i.yu@meta.com}
\author{Pace Nalbone}
\affiliation{%
  \institution{Codec Avatars Lab, Meta}
  \country{USA}}
\email{pacenalbone@meta.com}
\author{Matt Gramlich}
\affiliation{%
  \institution{Codec Avatars Lab, Meta}
  \country{USA}}
\email{matthewgramlich@meta.com}
\author{Roman Lubachersky}
\affiliation{%
  \institution{Codec Avatars Lab, Meta}
  \country{USA}}
\email{rlubachersky@meta.com}
\author{Chenglei Wu}
\affiliation{%
  \institution{Codec Avatars Lab, Meta}
  \country{USA}}
\email{chenglei@meta.com}
\author{Javier Romero}
\affiliation{%
  \institution{Codec Avatars Lab, Meta}
  \country{USA}}
\email{javierromero1@meta.com}
\author{Jason Saragih}
\affiliation{%
  \institution{Codec Avatars Lab, Meta}
  \country{USA}}
\email{jsaragih@meta.com}
\author{Michael Zollhoefer}
\affiliation{%
  \institution{Codec Avatars Lab, Meta}
  \country{USA}}
\email{zollhoefer@meta.com}
\author{Andreas Geiger}
\affiliation{%
  \institution{University of Tübingen}
  \country{Germany}}
\email{a.geiger@uni-tuebingen.de}
\author{Siyu Tang}
\affiliation{%
  \institution{ETH Zürich}
  \country{Switzerland}}
\email{siyu.tang@inf.ethz.ch}
\author{Shunsuke Saito}
\affiliation{%
  \institution{Codec Avatars Lab, Meta}
  \country{USA}}
\email{shunsuke.saito16@gmail.com}

\renewcommand\shortauthors{Wang, S. et al}

\begin{abstract}
We propose Relightable Full-Body Gaussian Codec Avatars, a new approach for modeling relightable full-body avatars with fine-grained details including face and hands.
The unique challenge for relighting full-body avatars lies in the large deformations caused by body articulation and the resulting impact on appearance caused by light transport.
Changes in body pose can dramatically change the orientation of body surfaces with respect to lights, resulting in both local appearance changes due to changes in local light transport functions, as well as non-local changes due to occlusion between body parts.
To address this, we decompose the light transport into local and non-local effects.
Local appearance changes are modeled using learnable zonal harmonics for diffuse radiance transfer.
Unlike spherical harmonics, zonal harmonics are highly efficient to rotate under articulation.
This allows us to learn diffuse radiance transfer in a local coordinate frame, which disentangles the local radiance transfer from the articulation of the body.
To account for non-local appearance changes, we introduce a shadow network that predicts shadows given precomputed incoming irradiance on a base mesh.
This facilitates the learning of non-local shadowing between the body parts.
Finally, we use a deferred shading approach to model specular radiance transfer and better capture reflections and highlights such as eye glints.
We demonstrate that our approach successfully models both the local and non-local light transport required for relightable full-body avatars, with a superior generalization ability under novel illumination conditions and unseen poses.
\end{abstract}

%
%


\begin{teaserfigure}
  \includegraphics[width=\textwidth]{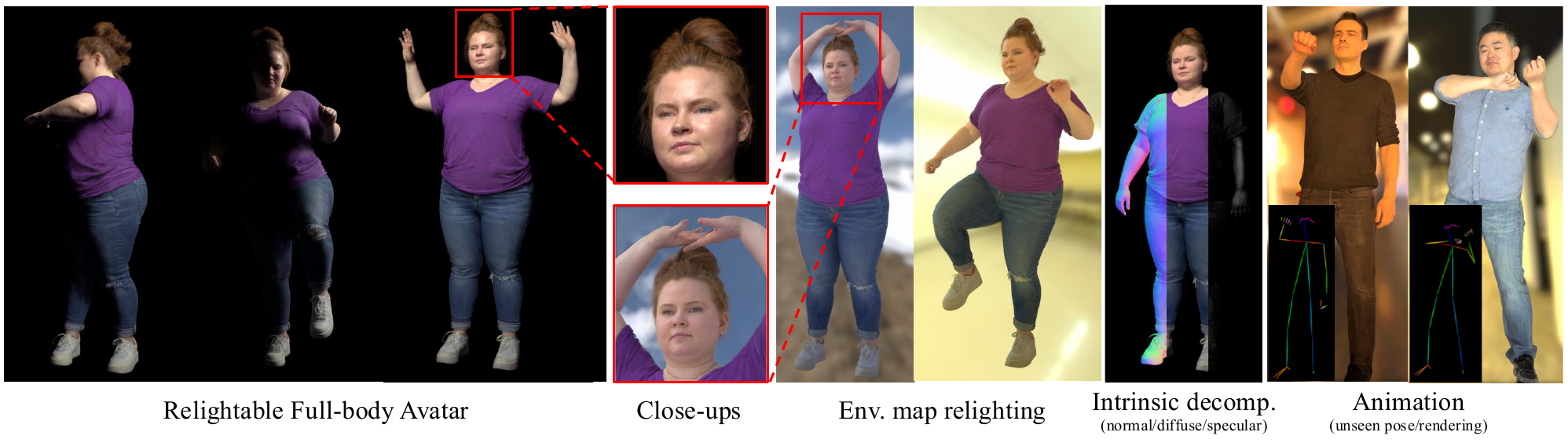}
  \vspace*{-6mm}
  \caption{\textbf{Relightable Full Body Gaussian Codec Avatars.}  We present
    the first approach that enables reconstruction, relighting and expressive
    animation of full-body avatars including body, face, and hands.  Our
    approach combines learned, orientation-dependent diffuse radiance transport
    and deferred-shading-based specular radiance transport to enable complex
    light transport such as global illumination for fully articulated human
    bodies.  } 
  \label{fig:teaser}
\end{teaserfigure}

\maketitle

\section{Introduction}
\label{sec:intro}
%
Building drivable full-body avatars is a long-standing challenge in computer vision and graphics.
Early approaches focused on reconstructing the geometry and appearance of the human body for free-viewpoint rendering and video playback~\cite{Collet2015SIGGRAPH,Prada2017SIGGRAPH,Starck2007CGA}.
While achieving high-fidelity appearance and geometry, these methods are limited in
their ability to animate the avatars under novel illumination conditions.
Later works recover intrinsic properties of the human body~\cite{Guo2019SIGGRAPH,Zhang2021SIGGRAPHa}, face~\cite{Bi2021SIGGRAPH,yang2023towards}, and hands~\cite{Iwase2023CVPR,Chen2024CVPR} to enable animation and relighting.
Among these approaches,~\cite{Guo2019SIGGRAPH,Zhang2021SIGGRAPHa,Chen2024CVPR} employ mesh-based representations, which often fail to model translucency and fine-scale geometric details such as hair.
~\cite{Iwase2023CVPR,yang2023towards} employ a mixture of volumetric primitives~\cite{lombardi2021mixture} that better captures fine-scale geometric detail compared to mesh-based representations, but tends to blur out certain geometric detail such as individual hair strands.
On the other hand, most religthtable appearance representations are also suboptimal:~\cite{Guo2019SIGGRAPH} employs a physically based rendering model that omits global illumination due to performance concern, thus producing unrealistic human skins.
~\cite{Zhang2021SIGGRAPHa,Bi2021SIGGRAPH,yang2023towards,Iwase2023CVPR,Chen2024CVPR} utilize neural relighting to predict relit appearance given the illumination as input.
These approaches can capture global illumination effects but often produce blurry appearance due to the limited expressiveness of the employed neural network.

Contrary to the aforementioned approaches,~\cite{Saito2024CVPR} explores 3D Gaussian Splatting (3DGS~\cite{Kerbl20233DGS}) to represent the geometry and appearance of head avatars and could represent highly detailed geometry such as individual hair strands.
The approach also employs a learnable radiance transfer function to account for global illumination effects.
The learned radiance transfer leverages spherical harmonics to model diffuse shading (~\cite{sloan2002precomputed}) and to capture low-frequency global illumination effects such as subsurface scattering of human skin.
In addition, specular radiance transfer is modeled based on a spherical Gaussian model~\cite{green2006view,wang2009all} to account for all-frequency illumination effects, such as eye glints and skin reflections.
Both components are directly compatible with conventional real-time rendering engines.

In this paper, we propose Relightable Full-Body Gaussian Codec Avatars, the first
approach to jointly model the relightable appearance of the body, face, and hands of drivable avatars.
We build upon the insights of~\cite{Saito2024CVPR}, using 3DGS as the underlying representation, while employing learned radiance transfer to model relightable appearance.
We note that there are several challenges to extend the appearance model of~\cite{Saito2024CVPR} to handle fully articulated bodies:
(1) the diffuse light transport model based on spherical harmonics in~\cite{Saito2024CVPR} assumes that light sources can be mapped to a single local coordinate frame (\ie, the head coordinate frame), which does not hold for articulated bodies, where each body part has its own local coordinate frame.
(2) Articulated bodies also exhibit complex shadowing effects caused by occlusions between body parts, which are not considered in~\cite{Saito2024CVPR}.
(3) Full-body models usually have a limited representational budget for modeling facial details compared to head-specific methods.
Naive splatting restricts resolution to the local Gaussian density, requiring many Gaussians for fine details. Moreover, because the resolution for specular reflections depends on both surface properties and the environment’s frequency content, modeling specularities at the Gaussian level forces an undesirable link between reflection frequency and Gaussian density, resulting in an under-representation of facial details such as eye glints.

To address the first challenge, we replace the diffuse light transport model based on spherical harmonic with zonal
harmonics~\cite{sloan2005local}, a representation that can be learned in the local coordinate frame and efficiently rotated to world coordinates, yielding distinct light transport effects for different body articulations with a single
parameterization.
In particular, zonal harmonics enable us to construct radiance transfer functions in world coordinates by efficiently rotating learned zonal harmonics parameters, circumventing the need to map light sources to the local coordinate frames of each body part.

Regarding shadow modeling, several recent full-body avatar works have proposed to use ray tracing to
account for shadowing effects~\cite{Chen2022ECCVa,Lin2024AAAI,Wang2024CVPR,Xu2024CVPR,Chen2024ECCV,Li2024ARXIV}.
They require expensive ray tracing of several rays per pixel at each optimization step in order to capture the shadows cast by intricate structures such as cloth wrinkles.
In contrast, our learned radiance transfer function captures local shadows well but
struggles with non-local shadows caused by distant self-occlusions.
We thus propose to learn a shadow network that is dedicated to predict the shadows caused by body articulation, given as input the normalized incoming irradiance on a coarse-tracked mesh.
The shadow network is inspired by~\cite{Bagautdinov2021DrivingsignalAF} but is adapted to the setting of a relightable
appearance model.
Specifically, we ensure that the irradiance is normalized in a physically based way, such that the learned shadow network 
generalizes to novel illumination conditions.
Lastly, to address the reduced quality in specular rendering, we take inspiration from deferred shading~\cite{Deering1988SIGGRAPH,thies2019deferred,gao2020deferred,Ye2024SIGGRAPH} and propose to model specular radiance transfer with deferred shading, which achieves high-fidelity specular reflections for the face region.

In summary, we make the following contributions:
\begin{itemize}
    \item We propose the first relightable full-body avatar model that jointly
        models the relightable appearance of the human body, face, and hands for
        high-fidelity relighting and animation.
    \item To handle full-body articulations with global light transport, we
        propose learnable zonal harmonics to represent local diffuse radiance
        transfer in the local coordinate frames of each Gaussian. This results
        in a reduced number of parameters and improved rendering quality
        compared to the commonly used spherical harmonics representation.
    \item We reformulate the learnable radiance transfer to explicitly decompose
        non-local shadowing, and propose a dedicated shadow network to predict
        shadows caused by the articulation of the body.  We also propose a
        physically based irradiance normalization scheme to ensure that the
        shadow network can generalize to novel illumination conditions such as
        unseen environment maps.
    \item We show that deferred shading can be used for our learned specular
        radiance transfer function. This achieves high-fidelity specular
        reflections for relightable human avatar modeling without excessively
        increasing the number of Gaussians.
\end{itemize}

\section{Related Work}
\label{sec:related}
\subsection{Full-Body Avatar Representations}
Mesh-based representations are popular because they provide a
native integration with existing graphics pipelines~\cite{Loper2015SMPLAS}.
Existing approaches for building mesh-based animatable avatar models use
pose- and latent-code conditioned neural networks to predict textures and
geometry deformations in UV
space~\cite{grigorev2019coordinate,Bagautdinov2021DrivingsignalAF,Xiang2022DressingA,Xiang2023DrivableAC}
or on top of graph-based representations~\cite{habermann2021real}.
More faithful reconstructions require more expressive representations than meshes.
Neural Radiance Fields (NeRF)~\cite{mildenhall2021nerf} have powered a number of
methods for neural rendering of human bodies~\cite{Peng2021CVPR, Liu2021SIGGRAPHASIA, Su2021NeurIPS, Weng2022CVPR, Wang2022ECCV, Li2022ECCV, Su2022ECCV, Jiang2022ECCV}.
These methods typically employ a NeRF conditioned on human motion, either in world or canonical space, by warping the rays with an articulated model for better generalization.
On the other hand, they are often limited by the slow training/inference speed of NeRF.
~\cite{Remelli2022SIGGRAPH,Chen2023NEURIPS} utilize an efficient variant of NeRF, 
\ie\ mixture of volumetric primitives~\cite{lombardi2021mixture} to enable both faithful reconstruction and real-time rendering.
Aside from NeRF, point-based representations~\cite{zheng2023pointavatar, Su2023NPCNP, Prokudin2023dynamic}
allow for more flexible topology modeling and exploit the notion of locality, which leads to more parameter-efficient models and better generalization.

Most recently, 3D Gaussian Splatting~\cite{Kerbl20233DGS} (3DGS) enabled both the high-performance of point-based representations 
and the expressiveness of radiance fields by modeling the scene with learnable Gaussian primitives.
3DGS has been extended to support dynamic scenes~\cite{Luiten2023Dynamic3G}, and subsequently several works introduced neural representations~\cite{Hu2024CVPR,Zielonka2023ARXIV,Li2024CVPR,Qian2024CVPR,Pang2024CVPR}
incorporating 3DGS-based appearance with articulated geometry priors to enable animatable full-body models.
\cite{Zielonka2023ARXIV} embeds Gaussian primitives into tetrahedral cages, as opposed to a commonly used linear blend skinning geometry proxy, with compositional payload produced by pose-conditioned MLPs.
~\cite{Li2024CVPR,Pang2024CVPR} parameterize the Gaussian primitives on a pre-defined UV texture space, and deploys a convolutional network in UV-space to decode highly detailed pose-dependent Gaussian appearance and deformations.
\cite{Hu2024CVPR,Qian2024CVPR} map a set of Gaussians - initialized with a SMPL~\cite{Loper2015SMPLAS} template in canonical space, using a standard linear blend skinning (LBS) model coupled with a learnable non-rigid deformation model.
In this work, we also build upon 3DGS due to its efficiency and expressiveness.
We note that most of the aforementioned methods focus on animation and novel view synthesis, 
while perceptually realistic relighting of full-body avatars is rarely explored in the literature, as
discussed in the next section.

\subsection{Avatar Relighting}
Recent portrait relighting methods ~\cite{sun2019single,pandey2021total,kim2024switchlight,ji2022geometry,kanamori2018relighting}
employ learning-based techniques operating in image space. 
\cite{sun2019single} uses an encoder-decoder neural network trained on light stage
data to regress the subject's appearance under novel illumination conditions.
\cite{kim2024switchlight} proposes an image-space approach that incorporates
physics-based decomposition and relies on self-supervised pre-training to
improve generalization from limited light-stage data.
\cite{He2024SIGGRAPH} employs diffusion
models~\cite{Ho2020NEURIPS,Song2021ICLRa,Song2021ICLRb,Rombach2022CVPR} to
predict relit images of human faces given conditioning face images and light
information.
Although promising, image-based techniques often produce geometrically and temporally
inconsistent results due to their limited ability to model 3D consistency.

Physically based rendering (PBR) techniques aim at estimating approximate
properties of the underlying material based on an approximate physics model.
Relightables~\cite{Guo2019SIGGRAPH} recover detailed intrinsic properties of
the human body from light-stage data using a mesh and PBR appearance model.
Relighting4D~\cite{Chen2022ECCVa} aims to obtain relightable avatars from
sparse-view or monocular videos with unknown light sources using a physically based
decomposition of the scene, where the neural fields produce normal, occlusion,
diffuse, and specular components rendered with a physically based
renderer. Later
works~\cite{Lin2024AAAI,Xu2024CVPR,Wang2024CVPR,Chen2024ECCV,Li2024ARXIV,Zheng2024ECCV} learn
such avatars in canonical spaces to facilitate animation while employing explicit
ray tracing to enhance the realism of relighting.
In general, PBR is not designed for efficient modeling of global illumination effects which are crucial
for rendering perceptually realistic images. Rendering global illumination effects with PBR requires multi-bounce path tracing which
is prohibitively slow for gradient-based optimization of dynamic avatar models.
~\cite{Zhang2021SIGGRAPHa,Bi2021SIGGRAPH,yang2023towards} propose to use neural relighting along with a 3D head model to achieve global illumination effects while being 3D consistent.
Neural relighting with shadow conditioning has also been explored for relightable hands~\cite{Iwase2023CVPR,Chen2024CVPR} exhibit more articulation compared to the human head, but their bottleneck-based neural relighting methods with mesh or mixture of volumetric primitives are unable to capture high-frequency specularities and geometric details as shown in~\cite{Saito2024CVPR}.
Contrary to all aforementioned methods, our approach utilizes a 3D-consistent representation~\cite{Kerbl20233DGS} with learnable radiance transfer functions~\cite{Saito2024CVPR} to model the relightable appearance. This ensures 3D-consistent and high-fidelity relighting of full-body avatars in an efficient manner, for both seen and unseen poses.

\subsection{Learned Radiance Transfer}
Modeling global illumination effects is a long-standing challenge in computer graphics~\cite{Pharr2023Book}. While PBR with Monte Carlo path tracing is the most accurate method for rendering global illumination effects, it is not amenable to real-time applications due to its high computational cost. To address this, precomputed radiance transfer (PRT)~\cite{sloan2002precomputed} has been proposed for real-time rendering of global illumination effects.
PRT approximates the light transport function using a set of compact basis functions such as
spherical harmonics (SH), which reduces shading computations to simple dot products between the SH coefficients of the illumination and the transfer coefficients.
Follow-up works have extended PRT to handle all frequency lighting~\cite{Ng2003SIGGRAPH,Tsai2006SIGGRAPH,green2006view,wang2009all} and learning via neural networks~\cite{Xu2022SIGGRAPH,Rainer2022CGF,Lyu2022ECCV}. Regarding dynamic
scenes such as dynamic human heads, both ~\cite{li2022eyenerf} and~\cite{Saito2024CVPR} learn diffuse light transport functions as sets of spherical harmonic coefficients.
We find that this representation is not sufficient to capture diffuse appearance changes due to full-body articulations.
Inspired by \cite{sloan2005local}, we choose Zonal Harmonics (ZH) to construct orientation-dependent light transport functions. Instead of aligning zonal harmonics with known SH coefficients as in~\cite{sloan2005local}, we propose to learn zonal harmonics directly from light stage data in an end-to-end manner, together with the other intrinsic properties. They can yield distinct light transport functions efficiently given different orientations of the primitives. This allows us to learn complex, orientation-dependent light transport for full-body
avatars from image observations only.

The learned view-dependent specular radiance transfer of~\cite{Saito2024CVPR} based on spherical Gaussians~\cite{wang2009all}, on the other hand, can be directly applied to full-body avatars by mapping camera viewing directions into local coordinate frames of 3D Gaussians. However, we observe that this approach performs poorly in highly specular regions when the number of Gaussians is limited. To address this, we propose to combine deferred shading with the learnable radiance transfer by rasterizing not only physically based properties (roughness and normals) but also light transport coefficients (visibility). While deferred shading has been explored with 3DGS~\cite{Ye2024SIGGRAPH}, we are the first to utilize it for learnable radiance transfer.

\section{Method}
\label{sec:method}
\begin{figure*}
\centering
  \includegraphics[width=1.0\textwidth]{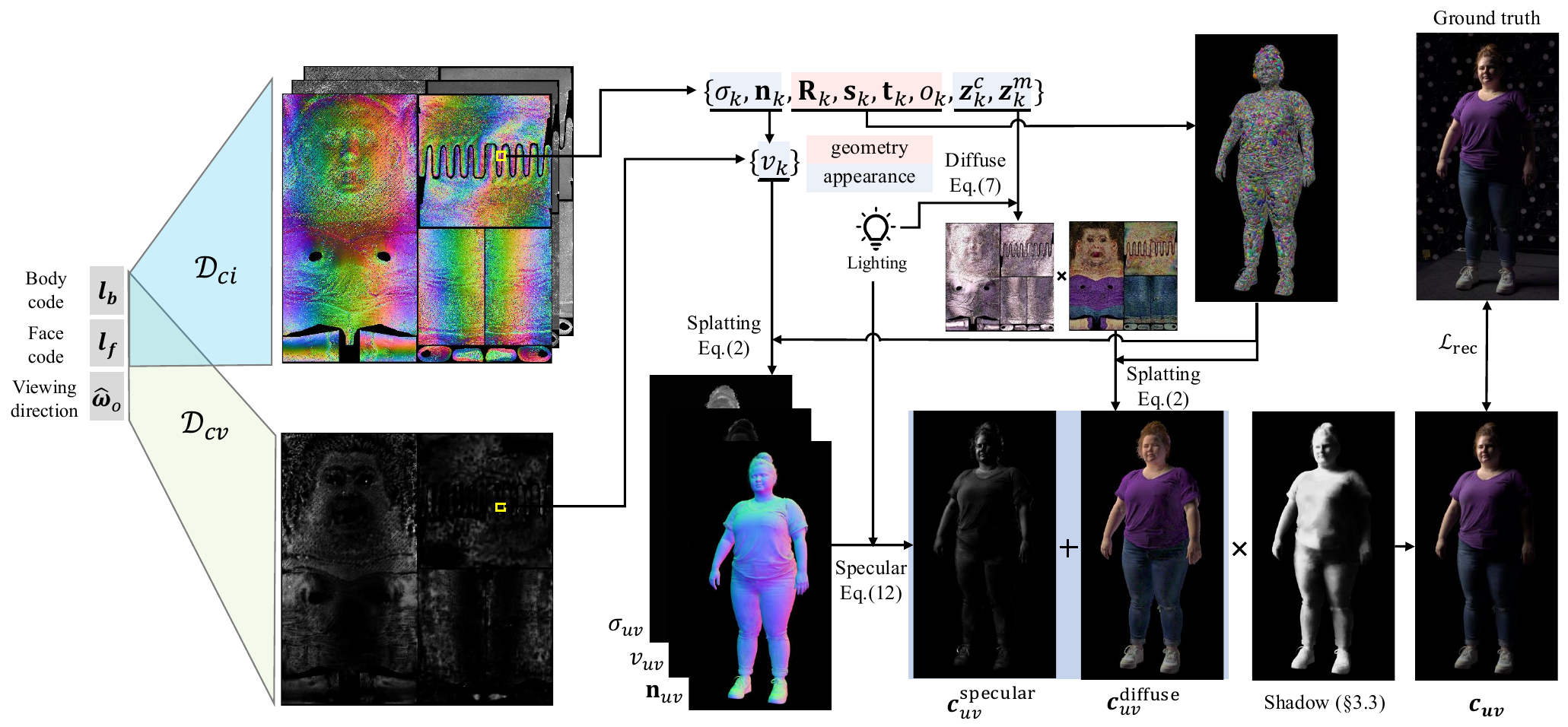}
    \caption{
    \textbf{Overview of our approach.}
    Given a body latent code $\mathbf{l}_b$ and a face latent code $\mathbf{l}_f$ computed by a keypoint encoder and canonicalized viewing directions $\hat{\mathbf{\omega}}_o$ as input, we decode the geometry parameters of 3D Gaussians $\{ \mathbf{R}_k, \mathbf{s}_k, \mathbf{t}_k, o_k \}$ (Sec.~\ref{sec:geometry}), and appearance parameters consisting of light transport coefficients $\{\mathbf{z}_k^c, \mathbf{z}_k^m\}$, normals $\{ \mathbf{n}_k \}$, roughness $\{\sigma_k \}$, and specular visibility $\{ v_k \}$ (Sec.~\ref{sec:appearance}).
    We integrate the light with diffuse light transport coefficients to yield per-Gaussian diffuse color, while using
    deferred shading to compute specular color.
    The final color is modulated by a shadow map predicted by a shadow network (Sec.~\ref{sec:shadowing}).
    }
  \label{fig:framework_RFGCA}
\end{figure*}
In this section, we describe in detail our method for relightable full-body
avatars as shown in Fig.~\ref{fig:framework_RFGCA}.

\subsection{Geometry}
\label{sec:geometry}
We represent full-body avatar as a collection of 3D Gaussians and employ 3D Gaussian splatting~\cite{Kerbl20233DGS} to render the avatar.
Similarly to~\cite{Saito2024CVPR}, we associate a Gaussian primitive with properties $\mathbf{g}_k = \{ \mathbf{t}_k, \mathbf{R}_k, \mathbf{s}_k, o_k, \rho_k, \mathbf{z}_k^c, \mathbf{z}_k^m, \mathbf{n}_k, v_k, \sigma_k \}$.
The geometry of the primitive is defined by a translation $\mathbf{t}_k \in \mathbb{R}^3$, a rotation $\mathbf{R}_k \in SO(3)$ represented as a quaternion, per-axis scales $\mathbf{s}_k \in \mathbb{R}^3$, and an opacity value $o_k \in [0, 1]$.
The appearance is defined by albedo $\rho_k \in \mathbb{R}^3$, diffuse light transport coefficients $\mathbf{z}_k^c, \mathbf{z}_k^m$, specular normal $\mathbf{n}_k \in \mathbb{S}^2$, specular visibility $v_k \in [0, 1]$, and
roughness $\sigma_k$.
The geometry of the $k$-th Gaussian primitive is modeled as an unnormalized 3D Gaussian kernel $\mathcal{G}_k$:
\begin{align}
    \label{eqn:gaussian_geometry}
    \mathcal{G}_k (\mathbf{x}) &= \exp \left( -\frac{1}{2} \left( \mathbf{x} - \mathbf{t}_k \right)^T \Sigma_k^{-1} \left( \mathbf{x} - \mathbf{t}_k \right) \right), \\
    \text{s.t.} \quad \Sigma_k &= \mathbf{R}_k \text{diag} (\mathbf{s}) \text{diag} (\mathbf{s})^T \mathbf{R}_k^T \nonumber
\end{align}
In order to render pixels in image space,~\cite{Kerbl20233DGS} uses an additional function $\mathcal{P} (\mathcal{G}_k, u, v)$ that projects the 3D Gaussian primitive onto the image plane~\cite{zwicker2002ewa}, and evaluates the Gaussian kernel value at the projected pixel location $(u, v)$.
The final color of a pixel is computed by blending the colors of all Gaussians, sorted by their
depth \wrt the camera:
\begin{align}
    \label{eqn:gaussian_splatting}
    \mathbf{C} (u, v) = \sum_{k=1} \mathbf{c}_k o_k \mathcal{P} (\mathcal{G}_k, u, v) \prod_{j=1}^{k-1} (1 - o_j \mathcal{P} (\mathcal{G}_j, u, v))
\end{align}
where $\mathbf{c}_k$ is the color of the $k$-th Gaussian.
Note that in our approach, we render the diffuse color with Eq.~\eqref{eqn:gaussian_splatting},
but use deferred shading for rendering specular color (Sec.~\ref{sec:specular_appearance}).
Similar to~\cite{Remelli2022SIGGRAPH,Bagautdinov2021DrivingsignalAF}, we parameterize rendering primitives (in our case, 3D Gaussians) on a UV texture map of a tracked human template mesh.
Given 3D body and face keypoints at a frame, we transform them according to the
inverse transformations of the body root and the face root, respectively, and
denote them as $\mathbf{K}_b, \mathbf{K}_f$.  We then encode the keypoints into
latent space and decode them into Gaussian primitives $\{ \mathbf{g}_k
\}_{k=1}^M$.
Formally, a encoder $\mathcal{E}$ and a view-independent decoder $\mathcal{D}_{ci}$ are defined as:
\begin{align}
    \label{eqn:ci_encoder_decoder}
    \mathbf{l}_b, \mathbf{l}_f &= \mathcal{E}(\mathbf{K}_b, \mathbf{K}_f; \Theta_e) \\
    \{\delta \mathbf{t}_k, \delta \mathbf{R}_k, \mathbf{s}_k, o_k, \mathbf{z}_k^c, \mathbf{z}_k^m, \delta \mathbf{n}_k\}_{k=1}^M &= \mathcal{D}_{ci}(\mathbf{l}_b, \mathbf{l}_f; \Theta_{ci})
\end{align}
where $\mathbf{l}_b$ and $\mathbf{l}_f$ are body and face latent codes predicted by the encoder.
The encoder $\mathcal{E}$ and the view-independent decoder $\mathcal{D}_{ci}$ are parameterized by
$\Theta_e$ and $\Theta_{ci}$, respectively.

In contrast to the face modeling approach of~\cite{Saito2024CVPR}, the human body exhibits a much
greater degree of articulation.
We thus propose to predict delta translation ($\delta \mathbf{t}_k$) and rotation ($\delta \mathbf{R}_k$) of each Gaussian primitive in a local coordinate frame, which is defined by the corresponding tangent-bitangent-normal (TBN) space of the tracked mesh.
Since each Gaussian is associated with a texel of the texture map, we have a TBN transformation for
each Gaussian primitive.
Let the TBN transformation for texel $k$ be $\mathbf{TBN}_k = [ \bar{\mathbf{t}}_k, \bar{\mathbf{b}}_k, \bar{\mathbf{n}}_k]$, where column vectors $\bar{\mathbf{t}}_k, \bar{\mathbf{b}}_k, \bar{\mathbf{n}}_k$ represent the tangent, bitangent, and normal at the texel $k$.
Let the 3D world coordinate of the texel $k$ be $\mathbf{v}_k$.
The translation and rotation of each Gaussian primitive in the world coordinate frame is then:
\begin{align}
    \label{eqn:geometry_transl}
    \mathbf{t}_k &= \mathbf{v}_k + \mathbf{TBN}_k \cdot \delta \mathbf{t}_k \\
    \label{eqn:geometry_rot}
    \mathbf{R}_k &= \mathbf{TBN}_k \cdot \delta \mathbf{R}_k 
\end{align}
where $\cdot$ denotes the matrix-matrix/matrix-vector multiplication.
We transform the quaternion $\delta \mathbf{R}_k$ to a rotation matrix before applying the TBN transformation, and then convert the resulting $\mathbf{R}_k$ back to a quaternion.

\subsection{Appearance}
\label{sec:appearance}
We follow the framework of~\cite{Saito2024CVPR} which models the relightable
appearance of a human face by combining diffuse light transport based on
spherical harmonics with a spherical-Gaussian-based specular light transport.
While~\cite{Saito2024CVPR} inversely maps incident light to the local coordinate
frame of head and compute diffuse shading in that local coordinate frame, it is
difficult to apply the same technique in the full-body scenario.  This is not
only because of the additional computational cost for mapping lights into local
coordinate frames of multiple body parts, but also because accurately modeling
inverse mappings for body joints is challenging.
It is thus preferable to rotate light transport functions to the world
coordinate, and compute diffuse shading in the world coordinate.

For specular light transport, we note that we cannot afford to use the same
number of Gaussian primitives for the face, compared to face-specific models.
This results in an under-representation of specular highlights in the face region.

In the following, we describe how to learn the diffuse transport coefficients in the local coordinate frame of each Gaussian primitive, which can be subsequently transformed to the world coordinate frame using the Gaussian rotation matrix.
We then describe a deferred shading scheme for specular light transport to improve the rendering quality of specular highlights.

\subsubsection{Zonal Harmonics for Diffuse Appearance}
\label{sec:zhs}
Following~\cite{Saito2024CVPR}, the diffuse color of the $k$-th Gaussian
primitive is defined as:
\begin{align}
    \label{eqn:diffuse_color}
    \mathbf{c}_k^d = \mathbf{\rho}_k \odot \int_{\mathbb{S}^2} \mathbf{L}(\mathbf{\omega}_i) \mathbf{d}_k (\mathbf{\omega}_i) \text{d} \mathbf{\omega}_i = \mathbf{\rho}_k \odot \sum_{i=1}^{(n+1)^2} \mathbf{L}_i \odot \mathbf{d}_k^i
\end{align}
in which $\mathbf{\omega}_i \in \mathbb{S}^2$ is the surface-to-light direction.
$\mathbf{L} = \{ \mathbf{L}_i \}_{i=1}^{(n+1)^2}$ and $\mathbf{d}_k = \{\mathbf{d}_k^i \}_{i=1}^{(n+1)^2}$ are the incident light and light transport coefficients represented as $n$-th order SH coefficients, respectively.
Both $\mathbf{L}_i$ and $\mathbf{d}_k^i$ are in $\mathbb{R}^3$.
$\mathbf{\rho}_k \in \mathbb{R}^3$ is the albedo for primitive $k$.
Albedos are defined and optimized directly on the UV texture map.
$\odot$ denotes the element-wise multiplication.

As discussed previously, we would like to rotate SH coefficients to the world coordinate
instead of mapping incident light to the local coordinate frames of body parts.
An immediate challenge is that rotating SH coefficients is prohibitively expensive,
especially for high-order SHs (we use $n=8$ in our experiments).
The amortized complexity of rotating SH coefficients is $O(n^3)$ for $n$th order SH.
To address this challenge, we take inspiration from~\cite{sloan2005local} and use Zonal Harmonics (ZHs) to model the
appearance of each Gaussian primitive in its local coordinate frame.
ZHs are a subset of SHs that are circularly symmetric around a specified direction.
In the simplest case, $\{\mathbf{d}_k^i \}_{i=1}^{(n+1)^2}$ can be represented as a function of arbitrary direction $\omega \in \mathbb{S}^2$, using a single set of ZH coefficients $\{ \mathbf{z}_k^l \}_{l=0}^{n}$:
\begin{align}
    \label{eqn:prt_zh}
    & \mathbf{d}^{i}_k (\omega) = \mathbf{z}^l_k Y_{lm} (\omega) \\
    \text{s.t.}& \quad \forall l = 0, \cdots, n, \quad \forall m = -l, \cdots, l \nonumber \\
    & \quad i = l^2 + l + m + 1 \nonumber
\end{align}
where $Y_{lm}$ is the SH basis function that maps a spherical direction onto the SH basis specified by $(l, m)$.
In this case, we predict only a single $\mathbf{z}^l_k$ for all $m$ values given a fixed $l$.
The ZH coefficients $\{\mathbf{z}^l_k \}_{l=0}^{n}$ are agnostic to the orientation of the primitive, which essentially represents the light transport properties of the primitive in a local coordinate frame.

Though efficient in yielding rotated SH coefficients, the expressiveness of a single ZH is limited in that Eq.~\eqref{eqn:prt_zh} can only represent functions that are circularly symmetric around $\omega$.
Thus in practice, we predict three sets of colored ZH coefficients, together denoted $\mathbf{z}_k \in \mathbb{R}^{3 \times 3l}$ for a texel $k$.
$\mathbf{d}_k$ is represented as the sum of these ZH basis functions evaluated at the tangent, bitangent, and normal directions of the Gaussian primitive, respectively:
\begin{align}
    \label{eqn:prt_zhs}
    \mathbf{d}_k^i &= \mathbf{z}_k^{0l} Y_{lm}(\hat{\mathbf{t}}_k) + \mathbf{z}_k^{1l} Y_{lm}(\hat{\mathbf{b}}_k) + \mathbf{z}_k^{2l} Y_{lm}(\hat{\mathbf{n}}_k)  \\ 
    \text{s.t.}& \quad \forall l = 0, \cdots, n, \quad \forall m = -l, \cdots, l \nonumber \\
    & \quad i = l^2 + l + m + 1 \nonumber
\end{align}
The tangent $\hat{\mathbf{t}}_k$, bitangent $\hat{\mathbf{b}}_k$, and normal $\hat{\mathbf{n}}_k$ directions are defined as the first, second, and third columns of $\mathbf{R}_k$ (Eq.~\eqref{eqn:geometry_rot}), respectively.
We represent colored ZHs ($\mathbf{z}_k^c$) up to the 3rd order while using monochromatic ZHs ($\mathbf{z}_k^m$) from the 4-th to 8-th order.

\subsubsection{Specular Appearance}
\label{sec:specular_appearance}
In this subsection, we describe how to model the specular appearance of the Gaussian primitives.
The general framework is similar to~\cite{Saito2024CVPR} but with modifications to adapt to full-body modeling.
We associate the specular normal vectors with the geometry of the Gaussian primitives, to obtain high-quality specular normals, especially for modeling clothes.
We also employ deferred shading to better capture specular highlights due to using a limited
number of Gaussians compared to face-only models.

\boldparagraph{Specular normal}
The normal vector $\mathbf{n}_k$ is crucial for modeling the specular appearance of the Gaussian primitive.
We found that associating the normal vector with the last column of the Gaussian primitive's rotation
matrix (\ie\ $\hat{\mathbf{n}}_k$ from Eq.~\eqref{eqn:prt_zhs}) achieves high-quality results.
Formally:
\begin{align}
\label{eqn:normal}
    \mathbf{n}_k = (\hat{\mathbf{n}}_k + \delta \mathbf{n}_k) / \| \hat{\mathbf{n}}_k + \delta \mathbf{n}_k \|_2
\end{align}
where $\delta \mathbf{n}_k$ is the predicted specular normal offset for the $k$-th Gaussian primitive.

\boldparagraph{Deferred shading for specular radiance transfer}
As demonstrated in previous works~\cite{Ye2024SIGGRAPH,Dihlmann2024NeurIPS}, deferred shading can also be applied to Gaussian splatting to improve the fidelity of the rendered specular appearance.
We employ a similar technique to our specular radiance transfer function.
We use Eq.~\eqref{eqn:gaussian_splatting} to render specular normals, roughness, and specular visibility in screen space, denoted as $\mathbf{n}_{uv}$, $\sigma_{uv}$, and $v_{uv}$, respectively.
Take the specular normal for example:
\begin{align}
    \label{eqn:deferred_normals}
    \bar{\mathbf{n}}_{uv} &= \sum_{k=1} \mathbf{n}_k o_k \mathcal{P} (\mathcal{G}_k, u, v) \prod_{j=1}^{k-1} (1 - o_j \mathcal{P} (\mathcal{G}_j, u, v))
\end{align}
The final screen space normal is defined as $\mathbf{n}_{uv} = \bar{\mathbf{n}}_{uv} / \| \bar{\mathbf{n}}_{uv} \|_2$.
$\sigma_{uv}$, and $v_{uv}$ are obtained similarly but without the normalization step.

\boldparagraph{Spherical Gaussians}
We employ spherical Gaussians~\cite{green2006view,wang2009all} to model the
specular appearance.  Given screen space parameters $\mathbf{n}_{uv}$,
$\sigma_{uv}$, and $v_{uv}$ which we have described in the previous section, we
compute the final specular color for the pixel $(u, v)$ in screen space as
follows:
\begin{align}
    \label{eqn:specular_color}
    \mathbf{c}^s (u, v) = v_{uv} \int_{\mathbb{S}^2} \mathbf{L}(\mathbf{\omega}_i) G_s(\mathbf{\omega}_i; \mathbf{q}_{uv}, \sigma_{uv}) \text{d} \mathbf{\omega}_i
\end{align}
where $G_s$ is the spherical Gaussian distribution of the specular lobe with mean $\mathbf{q}_{uv} \in \mathbb{S}^2$ and standard deviation $\sigma_{uv} \in \mathbb{R}^+ $.
Formally, the lobe is defined as:
\begin{align}
    \label{eqn:specular_gaussian}
    G_s(\mathbf{p}; \mathbf{q}_{uv}, \sigma_{uv}) = \frac{1}{\sqrt{2} \pi^{\frac{2}{3}} \sigma_{uv}} \exp \left( -\frac{1}{2} \left( \frac{ \arccos( \mathbf{p} \cdot \mathbf{q}_{uv} )}{\sigma_{uv}} \right)^2 \right)
\end{align}
in practice, the mean $\mathbf{q}_{uv}$ is the reflected vector of surface-to-camera direction around the surface normal: $\mathbf{q}_{uv} = 2 (\mathbf{n}_{uv} \cdot \mathbf{\omega}_o) \mathbf{n}_{uv} - \mathbf{\omega}_o$, where $\mathbf{\omega}_o$ is the surface-to-camera direction for pixel $(u, v)$.

\boldparagraph{View-dependent appearance decoder}
We decode specular parameters with a view-dependent decoder $\mathcal{D}_{cv}$:
\begin{align}
    \label{eqn:cv_decoder}
    \{ v_k \}_{k=1}^M &= \mathcal{D}_{cv}(\mathbf{l}_b, \mathbf{l}_f, \hat{\mathbf{\omega}}_o; \Theta_{cv})
\end{align}
where $\hat{\mathbf{\omega}}_o$ are canonicalized viewing directions in the local coordinate frames of the corresponding Gaussians.
Similar to the albedo, roughness $\sigma_k$ is defined explicitly on the UV texture
map and optimized with gradient descents.

\subsection{Learning shadowing effects}
\label{sec:shadowing}
Learning shadowing effects, especially for shadows caused by occlusion between body parts, is crucial for
realistic avatar appearance.
State-of-the-art methods rely on either mesh-based ray-tracing and denoising~\cite{Chen2024ECCV}, or tracing rays in radiance fields~\cite{Lin2024AAAI,Xu2024CVPR,Wang2024CVPR,Li2024ARXIV}.
The former is limited by the reconstruction quality of semi-opaque surfaces and structures, such as skin, hairs, and thin clothes.
The latter is limited by computational efficiency, as explicitly tracing rays in radiance fields is computationally expensive, and to estimate accurate shadowing effects, one needs to carry out ray tracing for each gradient update.
Fortunately, our learned radiance transfer model already captures local shadows caused by intricate geometry such as cloth wrinkles.
Here we describe the shadow branch that is dedicated to capturing non-local shadows caused by the occlusion between body parts.
We start by precomputing normalized irradiance for the underlying coarse tracked mesh $\mathbf{V} = \{ \mathbf{v}_k \}$ as follows:
\begin{align}
    \label{eqn:shadowing}
    \text{Irradiance} (\mathbf{v}_k) = \frac{\int_{\mathbb{S}^2} \mathbf{L}(\mathbf{v}_k, \mathbf{\omega}_i) \text{Vis} (\mathbf{v}_k, \mathbf{\omega}_i) \text{d} \mathbf{\omega}_i}{\int_{\mathbb{S}^2} \mathbf{L}(\mathbf{v}_k, \mathbf{\omega}_i) \text{d} \mathbf{\omega}_i}
\end{align}
where $\text{Vis} (\mathbf{v}_k, \mathbf{\omega}_i)$ is the visibility function that models whether the light from direction $\mathbf{\omega}_i$ is visible at $\mathbf{v}_k$.
We approximate Eq.~\eqref{eqn:shadowing} via Monte Carlo estimation in different scenarios such as multiple point lights (training) and environment maps (testing).
Details can be found in \iftoggle{arXiv}{Appendix~\ref{appx:mc_integration}}{the Appendix}.

We apply a light-weight convolutional neural
network~\cite{Bagautdinov2021DrivingsignalAF} in UV space to predict a shadow
map value $\text{shadow}_k \in [0, 1], \forall k \in \{ 1, \cdots, M\}$ given a
precomputed irradiance UV map.  Similar to specular normal, roughness, and
specular visibility, we render the shadow map in screen space as $\text{shadow}
(u, v)$.
The final output color for pixel $(u, v)$ is:
\begin{align}
    \label{eqn:final_color}
    \mathbf{C} (u, v) = (\mathbf{c}^d (u, v) + \mathbf{c}^s (u, v)) \cdot \text{shadow} (u, v)
\end{align}
where $\mathbf{c}^d (u, v)$ and $\mathbf{c}^s (u, v)$ are the diffuse and specular colors in screen space, respectively.

\subsection{Training Losses}
\label{sec:losses}
Given multi-view training videos of the target person along with the corresponding known illumination condition, we employ a standard L1 loss and LPIPS loss to supervise the reconstruction of the target person using the input RGB videos:
\begin{align}
    \mathcal{L}_{\text{rec}} = \mathcal{L}_{\text{L1}} + \lambda_{\text{LPIPS}} \mathcal{L}_{\text{LPIPS}}
\end{align}
where $\lambda_{\text{LPIPS}} = 0.1$.  In addition to the reconstruction loss, we also employ several regularization losses as follows:
\begin{align}
    \label{eqn:reg_loss}
    \mathcal{L}_{\text{reg}} =& \mathcal{L}_{\text{scale}} + \lambda_{\text{offset}} \mathcal{L}_{\text{offset}} + \lambda_{\text{mask}} \mathcal{L}_{\text{mask}} + \lambda_{\text{normal}} \mathcal{L}_{\text{normal}} \nonumber \\
    &+ \lambda_{\text{bound}} \mathcal{L}_{\text{bound}} + \lambda_{\text{normal\_orient}} \mathcal{L}_{\text{normal\_orient}} \nonumber \\
    &+ \lambda_{\text{alpha\_sparsity}} \mathcal{L}_{\text{alpha\_sparsity}} + \lambda_{\text{albedo}} \mathcal{L}_{\text{albedo}} \nonumber \\
    &+ \lambda_{\text{neg\_color}} \mathcal{L}_{\text{neg\_color}}
\end{align}
We refer readers to \iftoggle{arXiv}{Appendix~\ref{appx:loss_definition}}{the
Appendix} for a detailed definition of each loss term.

We optimize all trainable network parameters $\Theta = \{ \Theta_e, \Theta_{ci},
\Theta_{cv} \}$ and static parameters $\{ \rho_k, \sigma_k \}$ using Adam
optimizer.  We use a learning rate of $10^{-3}$ for network parameters while
$10^{-2}$ for static parameters.  Training runs for 300k iterations with
a batch size of 4 on a single NVIDIA A100 GPU, taking approximately 2 days.

\section{Experiments}
\label{sec:experiments}
In this section, we qualitatively and quantitatively evaluate our approach to building relightable full-body avatars.
We first summarize the dataset we captured for training and evaluation.
Then we introduce related baselines and evaluation metrics.
Finally, we present qualitative and quantitative results of our approach and the baselines, demonstrating the superior quality of our approach on the tasks of relighting and animating neural avatars.

\subsection{Dataset}
\label{sec:dataset}
\begin{figure}
    \centering
    \includegraphics[scale=0.5]{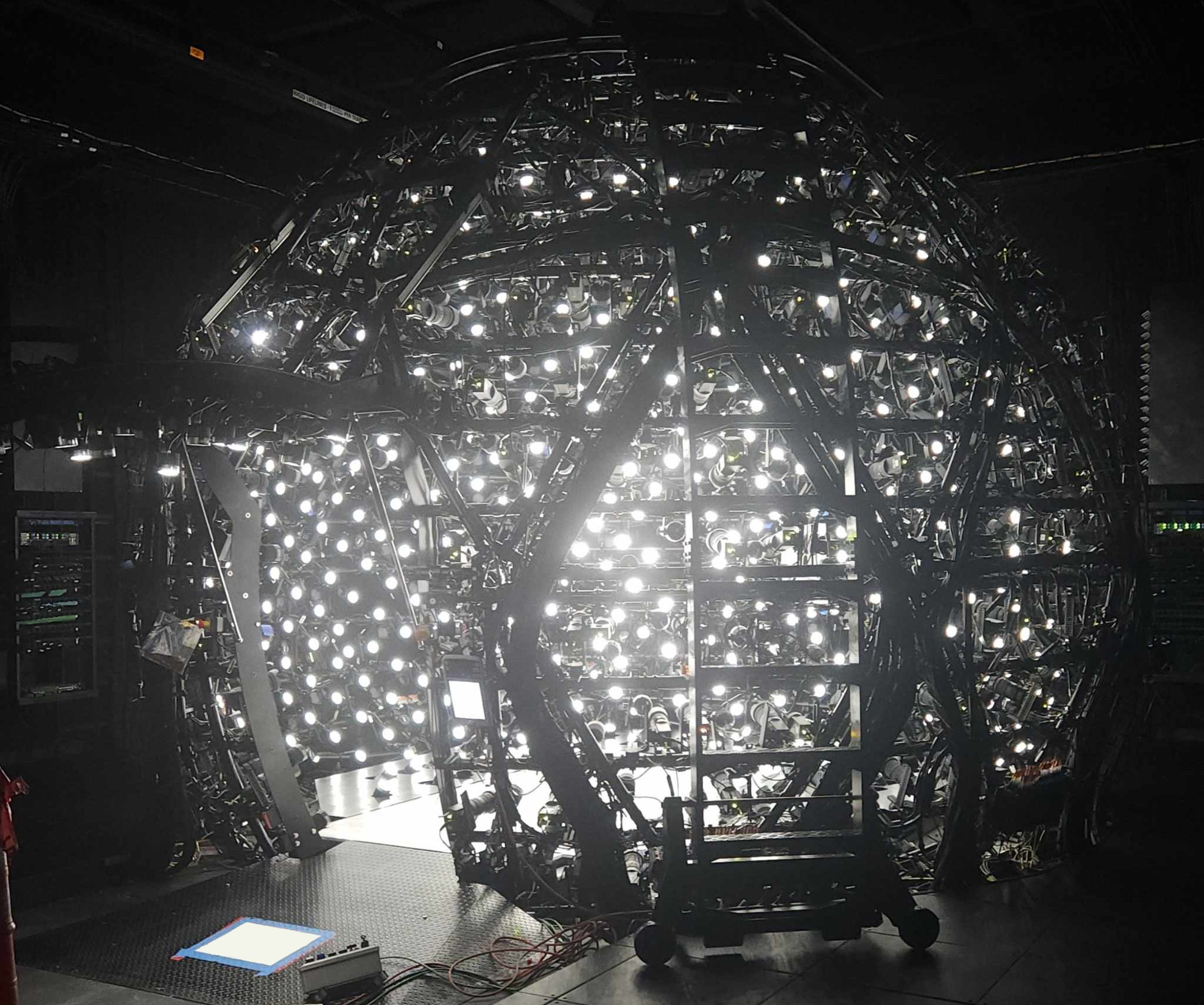}
    \caption{\textbf{Capture Dome.} Our multi-camera light stage with 512 cameras and 1024 controllable light sources. The dome has a radius of $2.75$ meters. Each camera has 24 mega-pixels resolution and records video with up to 90Hz.}
    \label{fig:lighticon}
\end{figure}

We captured five sequences using our multi-camera light stage, see Fig.~\ref{fig:lighticon}.
We employ three subjects for qualitative and quantitative evaluation against baselines, while the other two subjects are used to demonstrate additional qualitative results.
The light stage employs 1024 individually controllable light sources with known locations and light intensities.
The total number of training frames for each captured sequence is about 5000-6000, with 512 cameras for each frame.
The resolution of the captured videos is $5328x4608$.
We down-sample the capture to quarter resolution for more efficient training.
The captured videos consist of fully-lit frames (all light sources are on) and partially-lit frames (a random subset of 10-20 light sources are on).
We hold out 20\% of the camera views for evaluation.
We also hold out 10\% of the partially-lit frames from the training sequences as well as partially-lit frames from unseen motion sequences for evaluation.

\subsection{Baselines and Evaluation Metrics}
\label{sec:baselines}
\boldparagraph{Baselines}
Since there is no existing method that can directly run on our dataset (hundreds of high-resolution cameras, with calibrated and known light sources), we create a baseline that uses the learned geometry from
our method and a PBR appearance model that is employed in most established full-body avatar methods, \eg\
~\cite{Chen2022ECCVa,Xu2024CVPR,Lin2024AAAI,Wang2024CVPR,Chen2024ECCV,Li2024ARXIV}.
For ablations, we demonstrate the effectiveness of the ZH diffuse appearance representation and the importance of non-local shadow modeling.
We also show that associating Gaussian rotations with specular normals results in more detailed normal estimations, while deferred shading helps to capture detailed specular reflections such as eye
glints.

\boldparagraph{Evaluation Tasks and Metrics}
We quantitatively evaluate the performance of our method as well as baselines on the task of relighting using held-out poses from novel views.

We use standard PNSR/SSMI/LPIPS metrics for evaluation.  We also crop out the foreground human avatar before computing these metrics to minimize the influence from the background.

\subsection{Results and Discussion}
\begin{figure}
    \centering
    \begin{subfigure}[b]{0.31\linewidth}
        \centering
        \includegraphics[width=\linewidth, trim=10cm 8cm 25cm 4cm, clip]{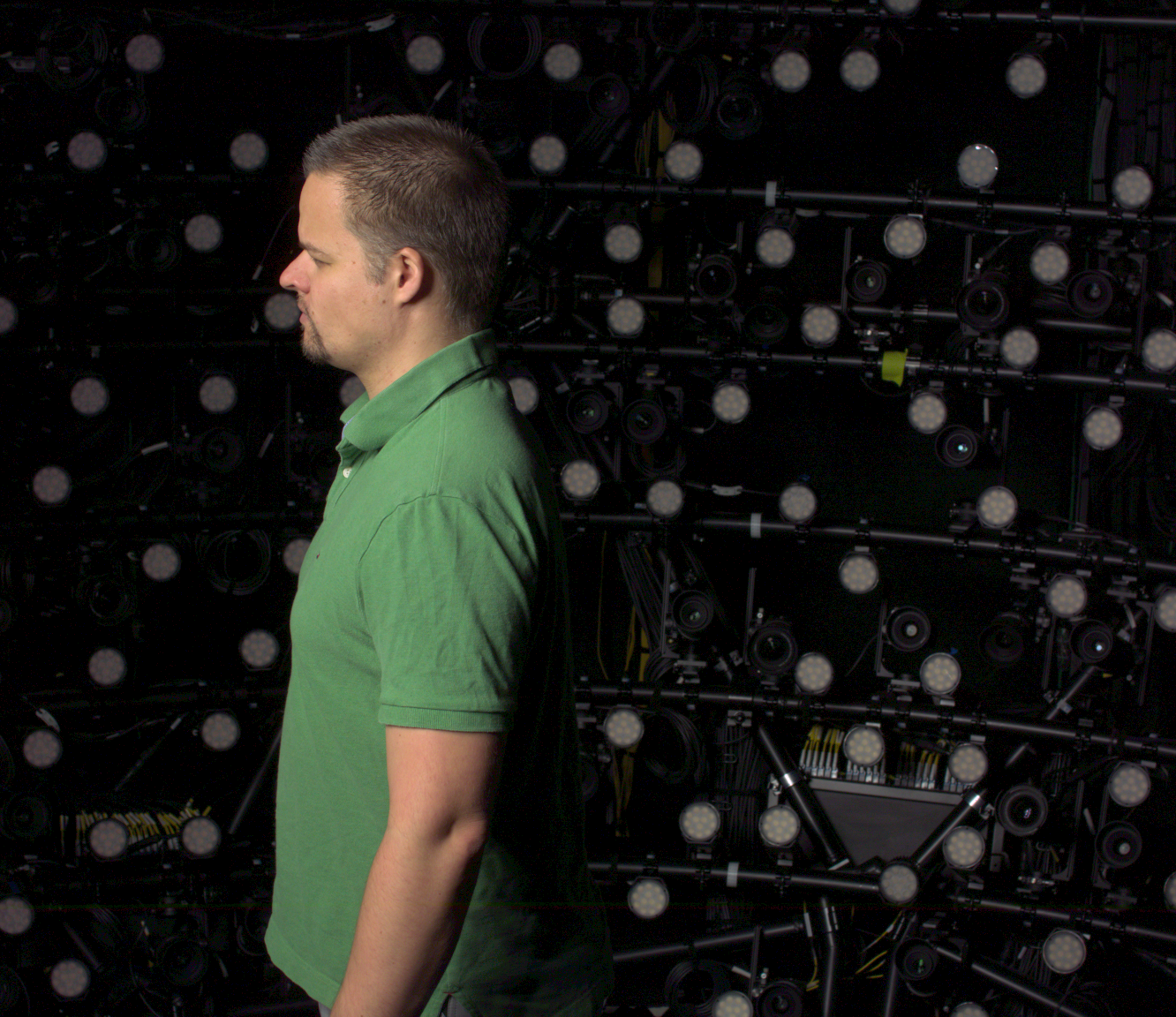}
        \caption{GT}
    \end{subfigure}
    \begin{subfigure}[b]{0.31\linewidth}
        \centering
        \includegraphics[width=\linewidth, trim=10cm 8cm 25cm 4cm, clip]{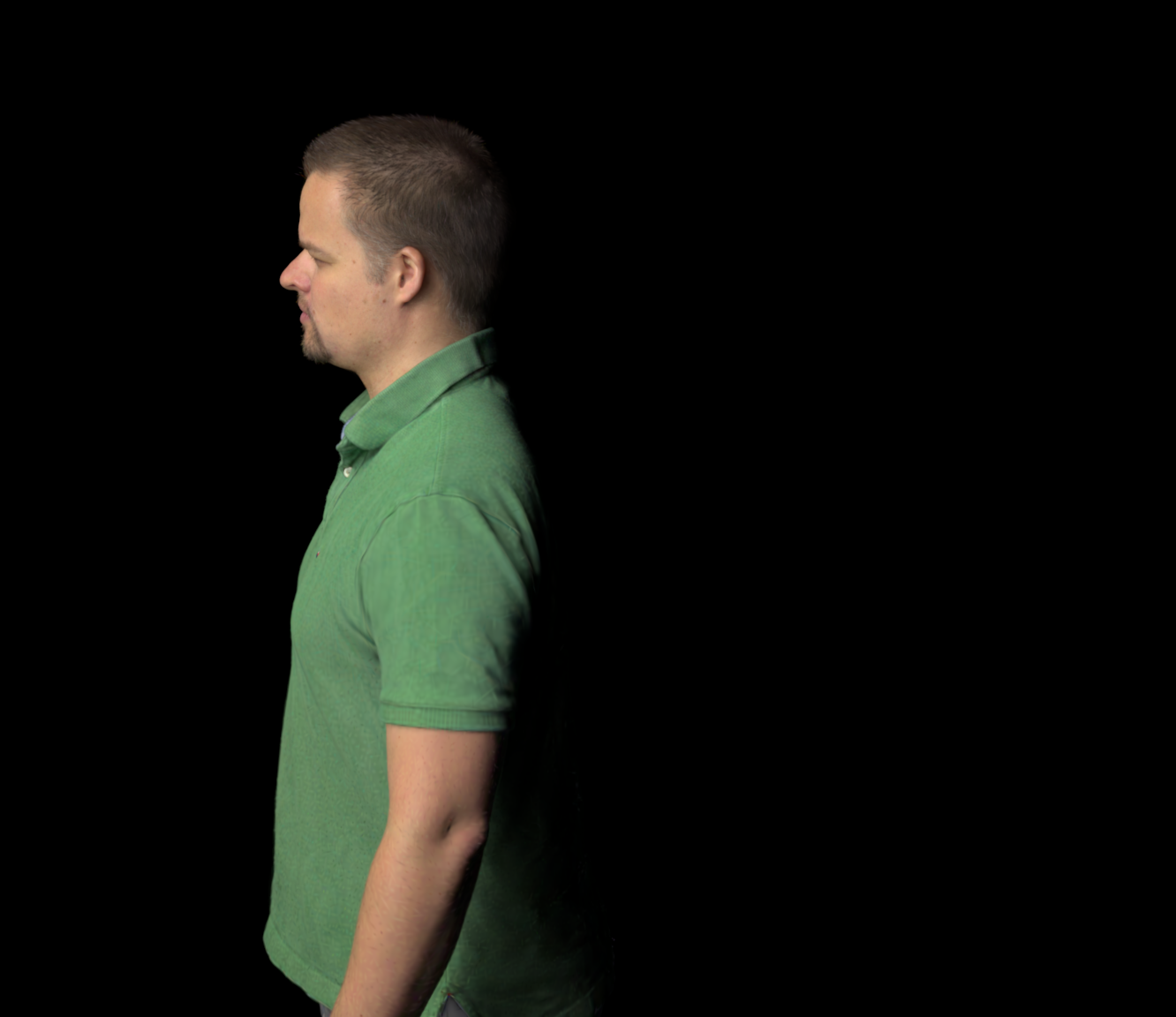}
        \caption{Ours}
    \end{subfigure}
    \begin{subfigure}[b]{0.31\linewidth}
        \centering
        \includegraphics[width=\linewidth, trim=10cm 8cm 25cm 4cm, clip]{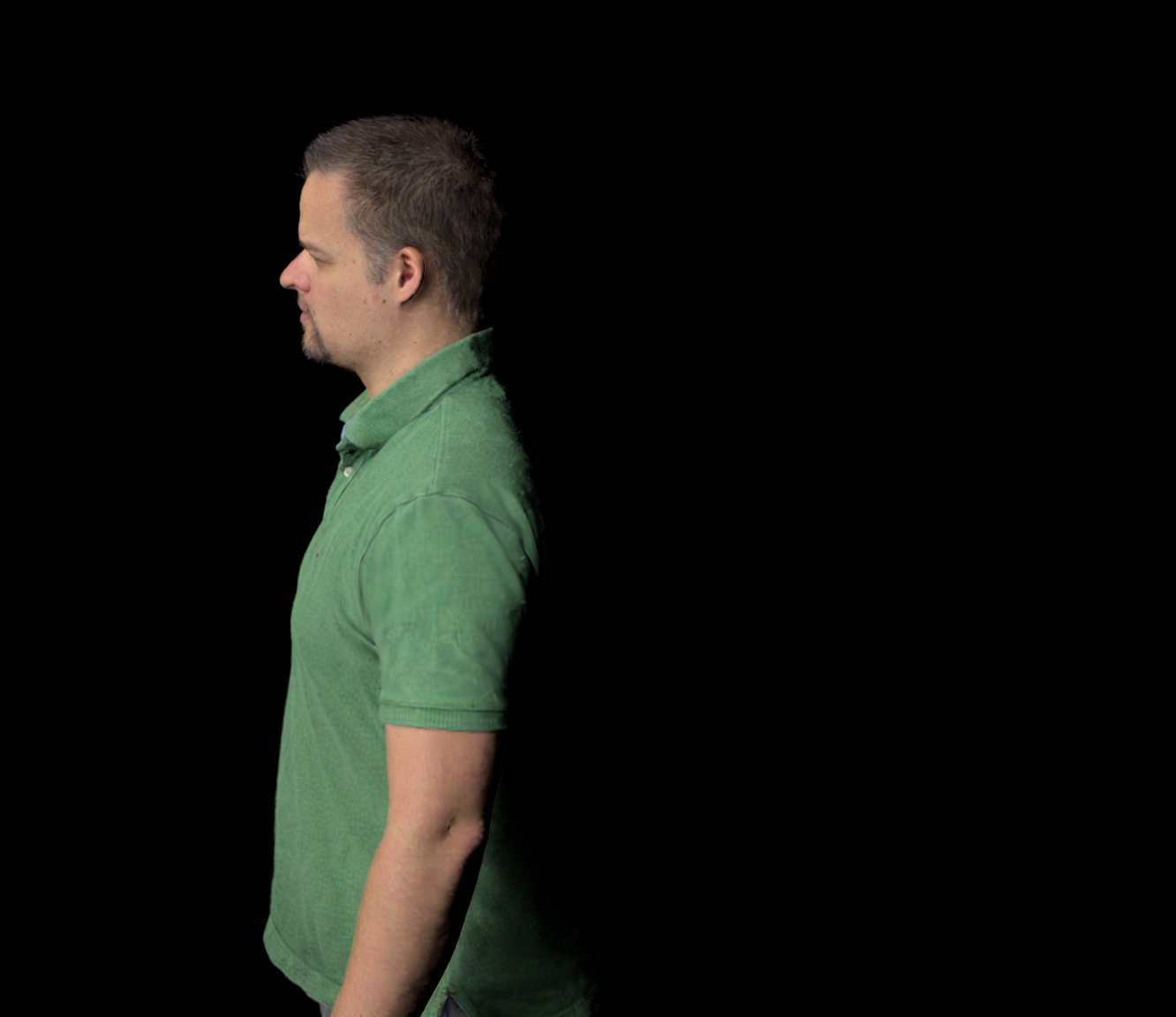}
        \caption{PBR}
    \end{subfigure}
    \caption{\textbf{Our appearance model vs. PBR appearance model.}  The PBR
    appearance model fails to capture subsurface scattering effects for skins
    and translucent structures such as hairs.  It also produces a darker appearance
    for concave structures such as ears by omitting global illumination.}
    \label{fig:qualitative_pbr}
\end{figure}

\begin{figure}
    \centering
    \begin{subfigure}[b]{0.31\linewidth}
        \centering
        \includegraphics[width=\linewidth, trim=4cm 5cm 15cm 5cm, clip]{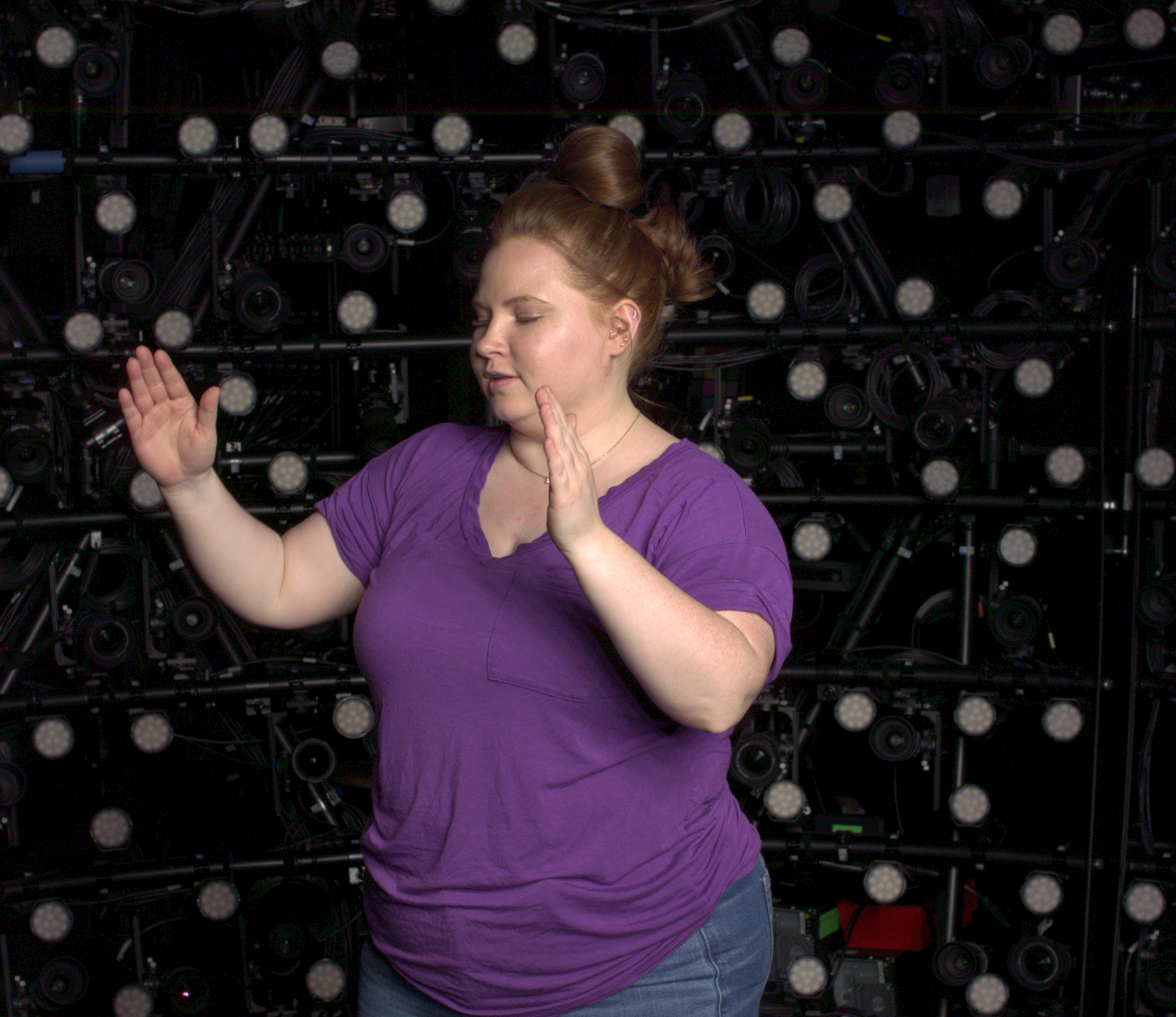}
        \caption{GT}
    \end{subfigure}
    \begin{subfigure}[b]{0.31\linewidth}
        \centering
        \includegraphics[width=\linewidth, trim=4cm 5cm 15cm 5cm, clip]{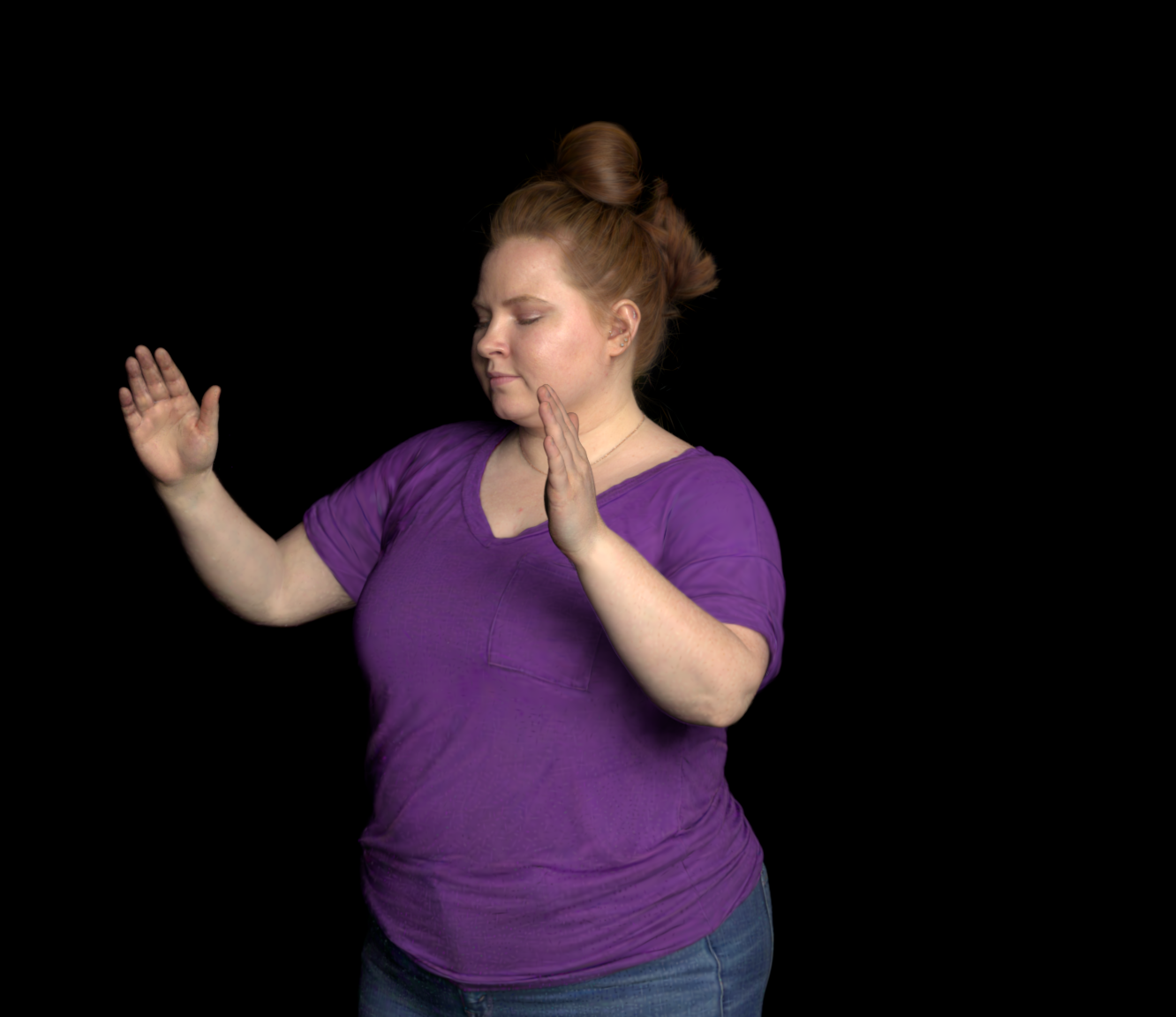}
        \caption{Ours (ZH)}
    \end{subfigure}
    \begin{subfigure}[b]{0.31\linewidth}
        \centering
        \includegraphics[width=\linewidth, trim=4cm 5cm 15cm 5cm, clip]{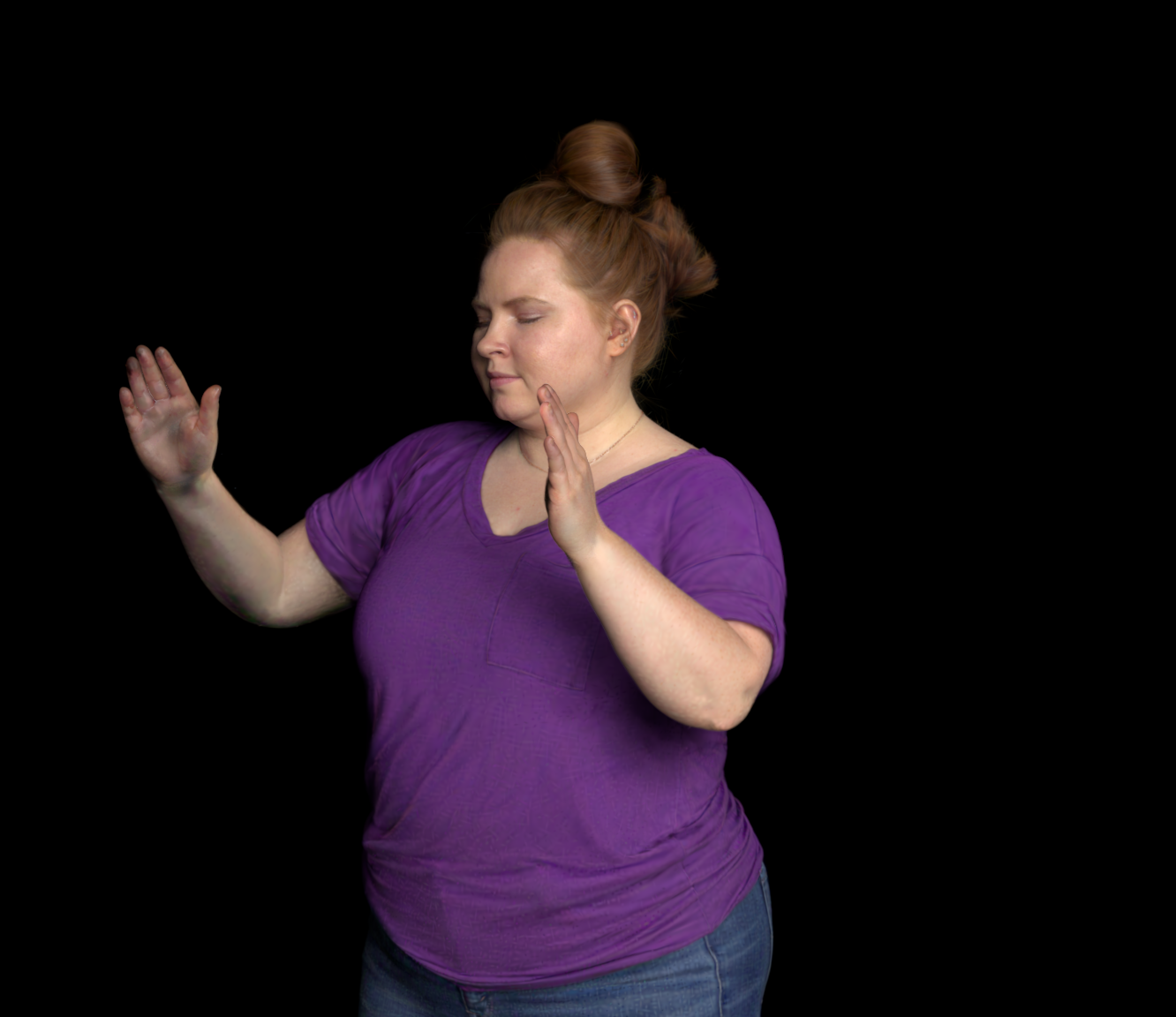}
        \caption{SH}
    \end{subfigure}
    \caption{\textbf{ZH vs. SH for diffuse light transport.} Note the incorrect
    shading on the right arm in the SH variant.}
    \label{fig:qualitative_sh}
\end{figure}

\begin{figure}
    \centering
    \begin{subfigure}[b]{0.31\linewidth}
        \centering
        \includegraphics[width=\linewidth, trim=4cm 5cm 15cm 5cm, clip]{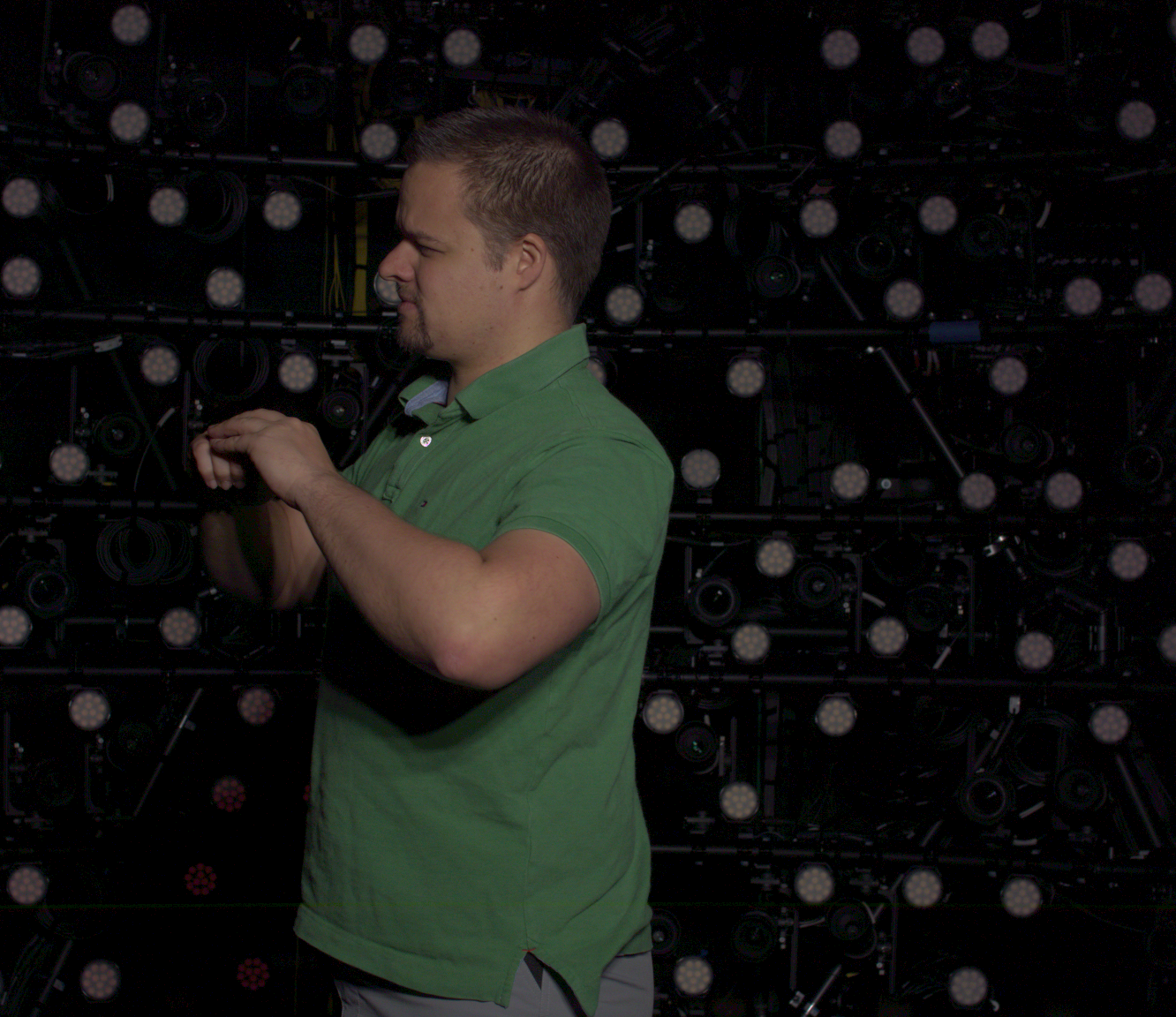}
        \caption{GT}
    \end{subfigure}
    \begin{subfigure}[b]{0.31\linewidth}
        \centering
        \includegraphics[width=\linewidth, trim=4cm 5cm 15cm 5cm, clip]{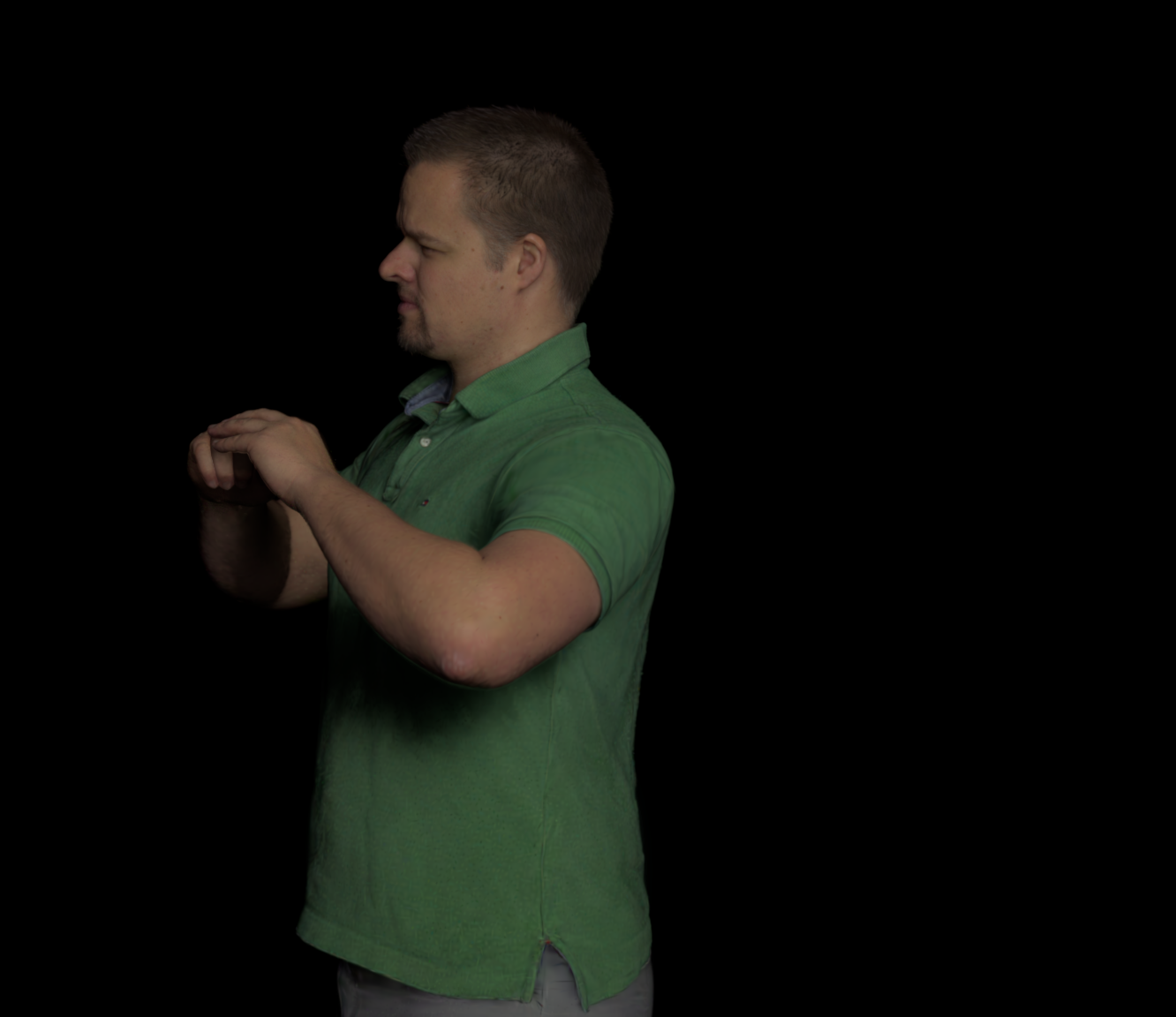}
        \caption{Ours (w. shadow)}
    \end{subfigure}
    \begin{subfigure}[b]{0.31\linewidth}
        \centering
        \includegraphics[width=\linewidth, trim=4cm 5cm 15cm 5cm, clip]{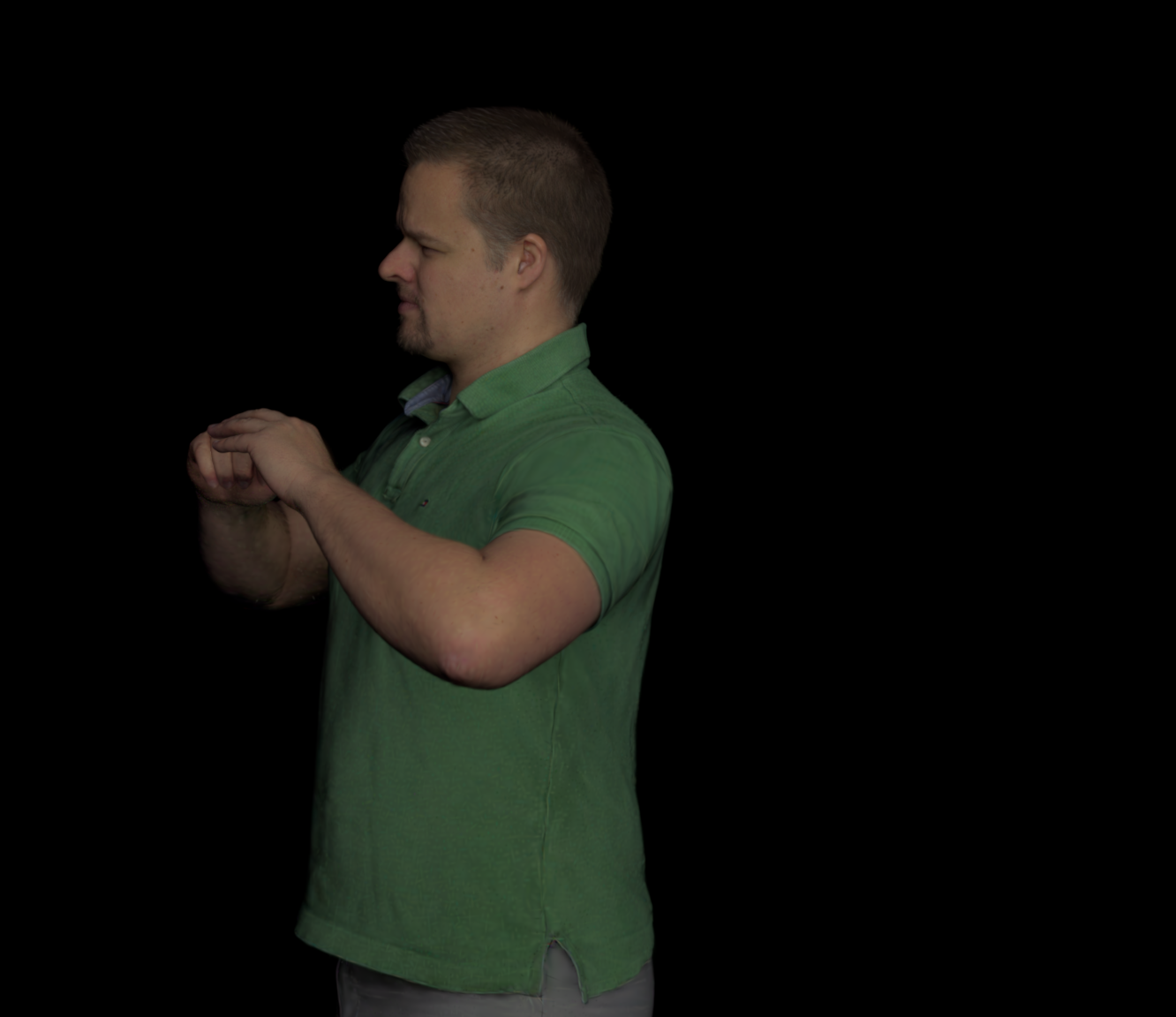}
        \caption{w.o. shadow}
    \end{subfigure}
    \caption{\textbf{Qualitative results shadow networks.} The learned light
    transport is not sufficient to capture the shadowing effects
    caused by body articulation without the help of the shadow network.}
    \label{fig:qualitative_shadow}
\end{figure}

\begin{figure}
    \centering
    \begin{subfigure}[b]{0.31\linewidth}
        \centering
        \includegraphics[width=\linewidth, trim=10cm 0cm 10cm 10cm, clip]{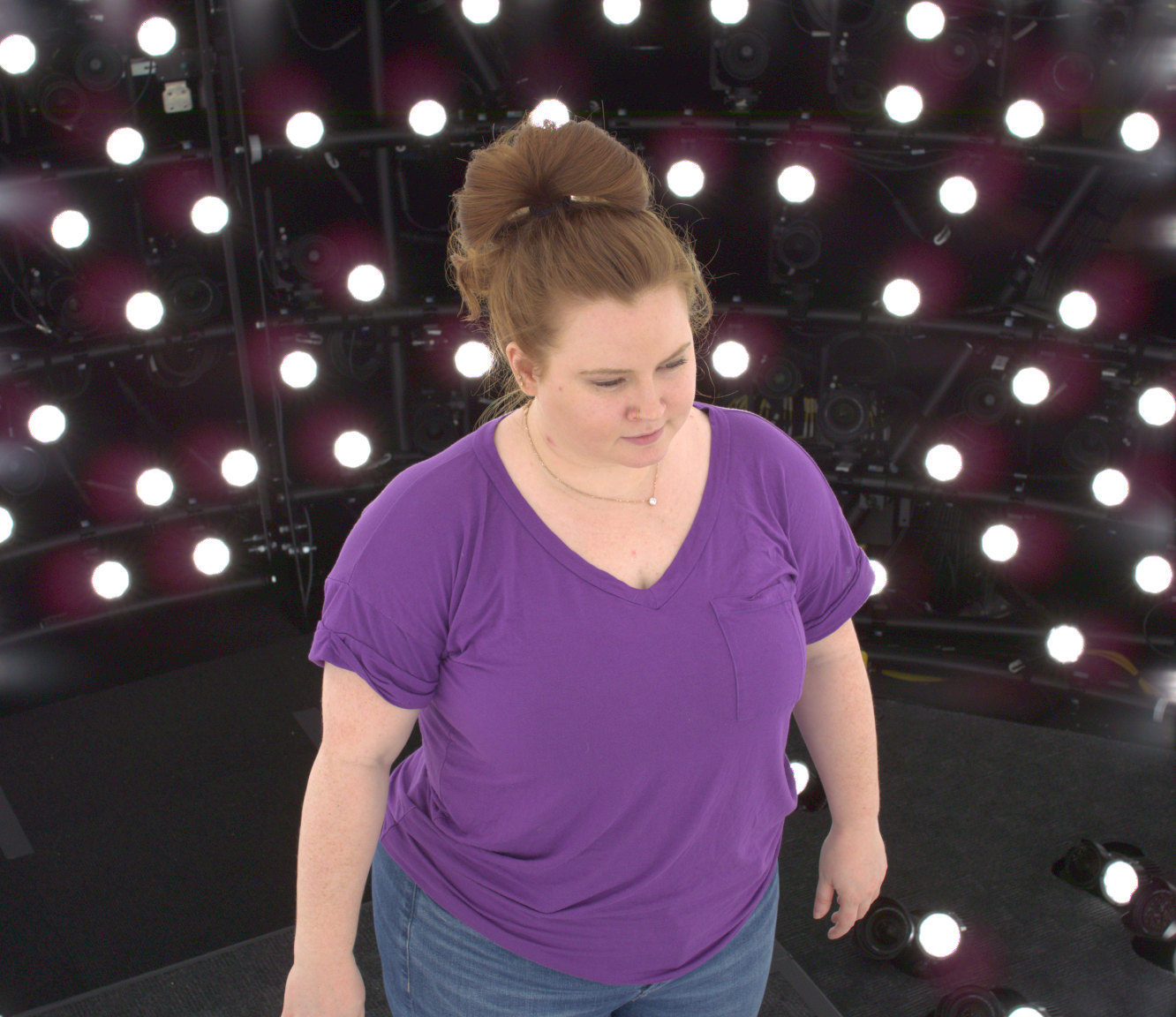}
        \caption{Reference}
    \end{subfigure}
    \begin{subfigure}[b]{0.31\linewidth}
        \centering
        \includegraphics[width=\linewidth, trim=10cm 0cm 10cm 10cm, clip]{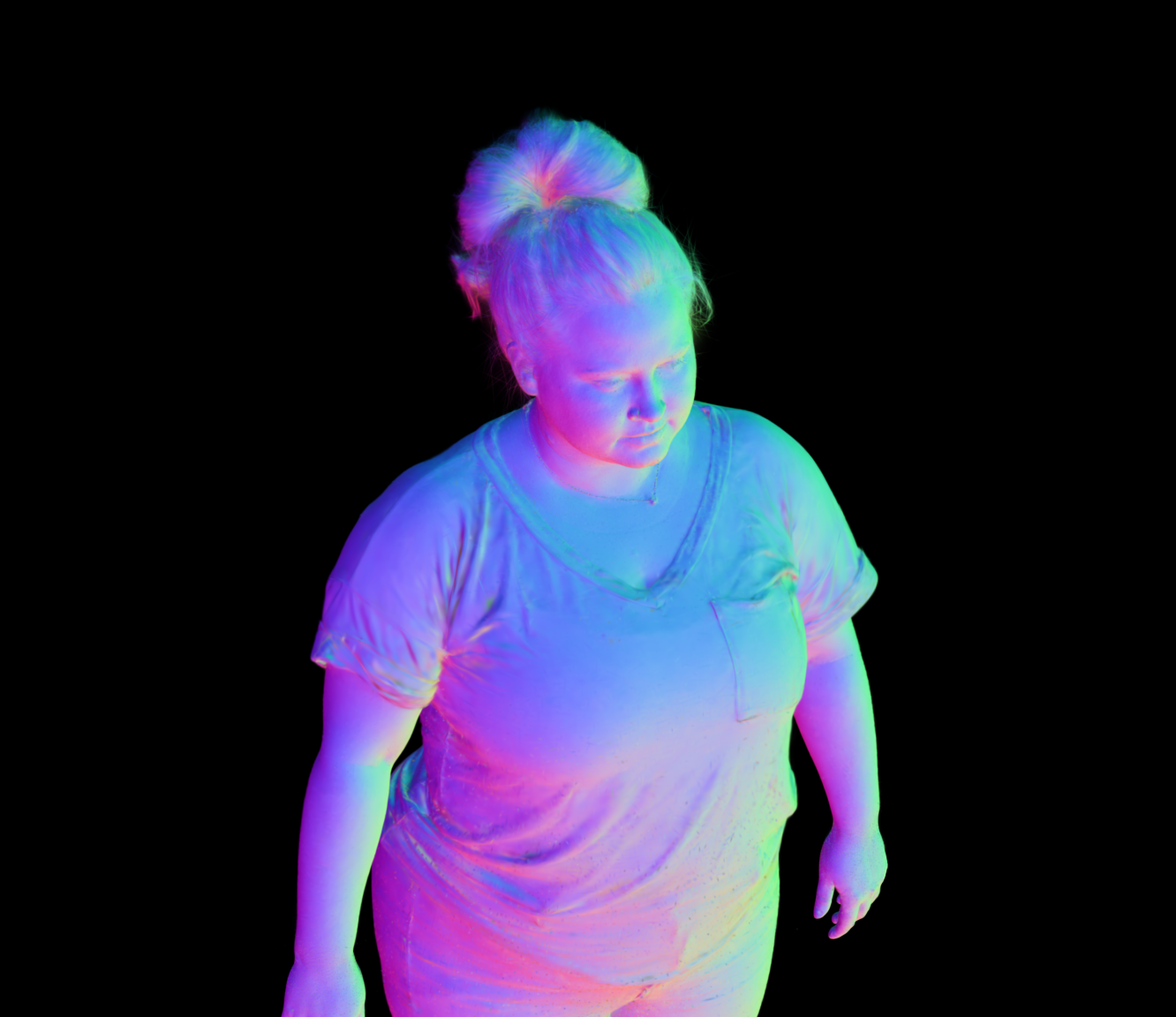}
        \caption{Ours}
    \end{subfigure}
    \begin{subfigure}[b]{0.31\linewidth}
        \centering
        \includegraphics[width=\linewidth, trim=10cm 0cm 10cm 10cm, clip]{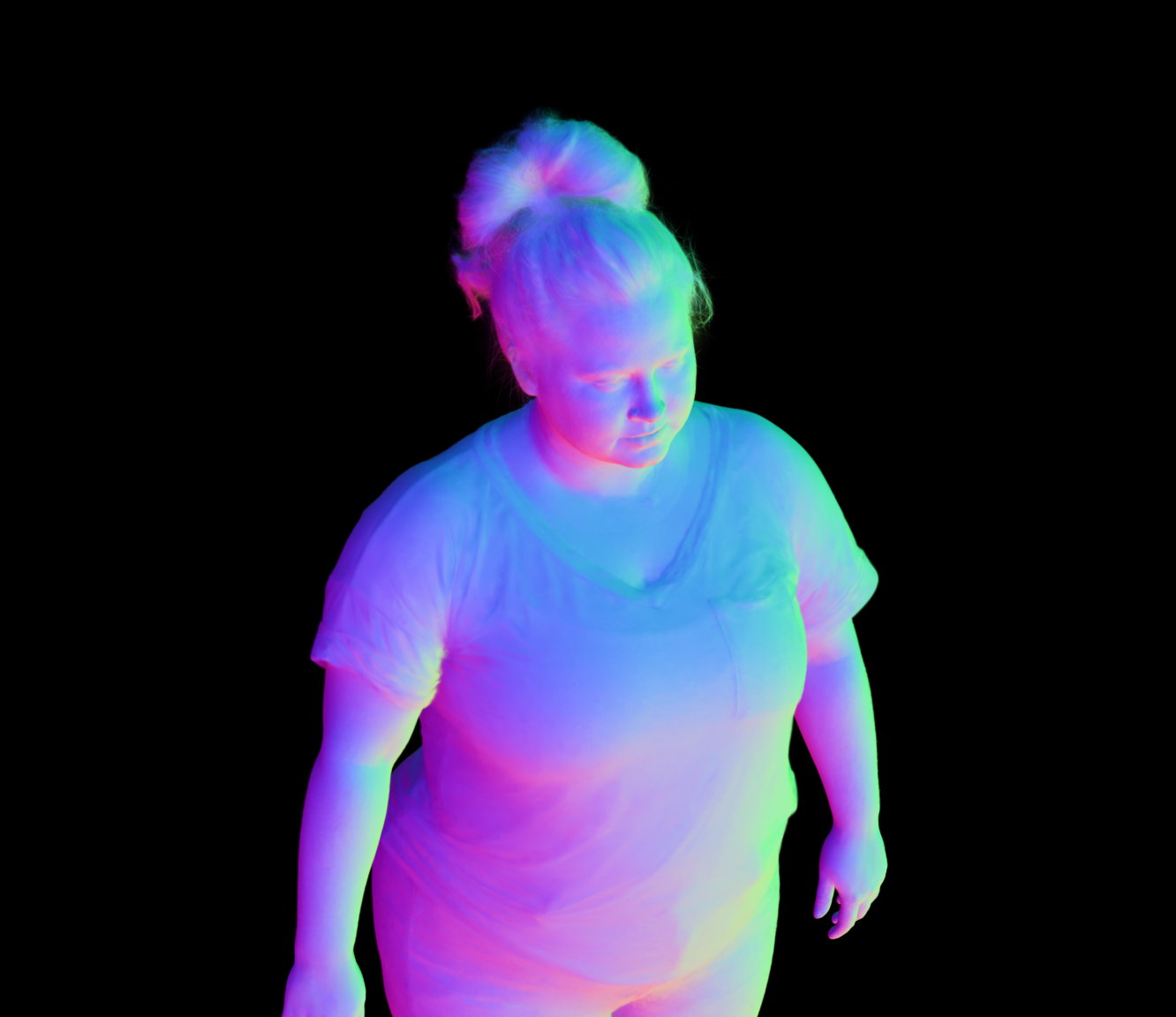}
        \caption{w. mesh normal}
    \end{subfigure}
    \caption{\textbf{Normal representations.}  
    The quality of normal estimation is significantly improved if Gaussian rotations
    are associated with specular normals.}
    \label{fig:qualitative_normal}
\end{figure}

\begin{table}
\small
\renewcommand{\tabcolsep}{2.0pt}
    \centering
    \begin{tabular}{l|c|c|c|c|c|c|}
        \toprule
        \multirow{2}{*}{Method} & \multicolumn{3}{c|}{Training Motion} & \multicolumn{3}{c}{Unseen Motion} \\
        \cmidrule{2-7}
        & PSNR $\uparrow$ & SSIM $\uparrow$ & LPIPS $\downarrow$ & PSNR $\uparrow$ & SSIM $\uparrow$ & LPIPS $\downarrow$ \\
        \midrule
        PBR            & 28.35 & 0.7729 & 0.1993 & 26.83 & 0.7477 & 0.2166 \\
        \midrule
        SH             & 29.15 & 0.7958 & 0.1846 & 27.21 & 0.7679 & 0.2056 \\
        w.o. shadow    & 28.89 & 0.7991 & 0.1800 & 27.07 & 0.7707 & 0.2004 \\
        w.o. deferred  & \cellcolor{rred} 29.55 & \cellcolor{rred} 0.8047 & 0.1796 & \cellcolor{rred} 27.59 & \cellcolor{oorange} 0.7755 & 0.2003 \\
        Mesh normal    & 29.43 & 0.8036 & \cellcolor{oorange} 0.1785 & 27.53 & 0.7747 & \cellcolor{oorange} 0.1993 \\
        \midrule
        Ours           & \cellcolor{oorange} 29.48 & \cellcolor{oorange} 0.8046 & \cellcolor{rred} 0.1781 & \cellcolor{oorange} 27.57 &\cellcolor{rred}  0.7756 & \cellcolor{rred} 0.1989 \\
        \bottomrule
    \end{tabular}
    \caption{\textbf{Quantitative comparison to baselines.}  The top two 
 approaches are highlighted in \textcolor{rred}{red} and \textcolor{oorange}{orange}, respectively.}
    \label{tab:quantitative}
\end{table}
We report the quantitative results in Table~\ref{tab:quantitative}.
Our learned radiance transfer model significantly outperforms the PBR appearance model in terms of all metrics.
This is because the PBR appearance model used in previous methods is designed mostly for opaque objects and does not model translucent structures such as hairs, and subsurface scattering effects for skins (Fig.~\ref{fig:qualitative_pbr}).
Our method also achieves the best LPIPS scores compared to all ablation variants.
Specifically, we show a large performance drop when using SH instead of ZH, which demonstrates the importance of the ZH diffuse coefficients that capture appearance more faithfully for highly articulated body parts such as hands and arms (Fig.~\ref{fig:qualitative_sh}).  Here SH is not rotated
as discussed int Sec.~\ref{sec:zhs}
We also note that the SH representations need $3 \times (3 + 1)^2 + (8 + 1)^2 - (3 + 1)^2 = 113$ parameters to represent a texel, whereas our ZH representation only needs $3 \times 3 \times (3 + 1) + 3
\times 5 = 51$ parameters.
Removing the shadow network also leads to a noticeable decrease in all metrics, indicating that a naive pose-dependent radiance transfer model is not sufficient to capture non-local shadow effects (Fig.~\ref{fig:qualitative_shadow}).
Replacing Gaussian normals with mesh normals also results in less detailed normals (Fig.~\ref{fig:qualitative_normal}),
and a slight drop in all metrics.

\begin{figure}
    \centering
    \begin{subfigure}[b]{0.31\linewidth}
        \centering
        \includegraphics[width=\linewidth, trim=13cm 0cm 10cm 20cm, clip]{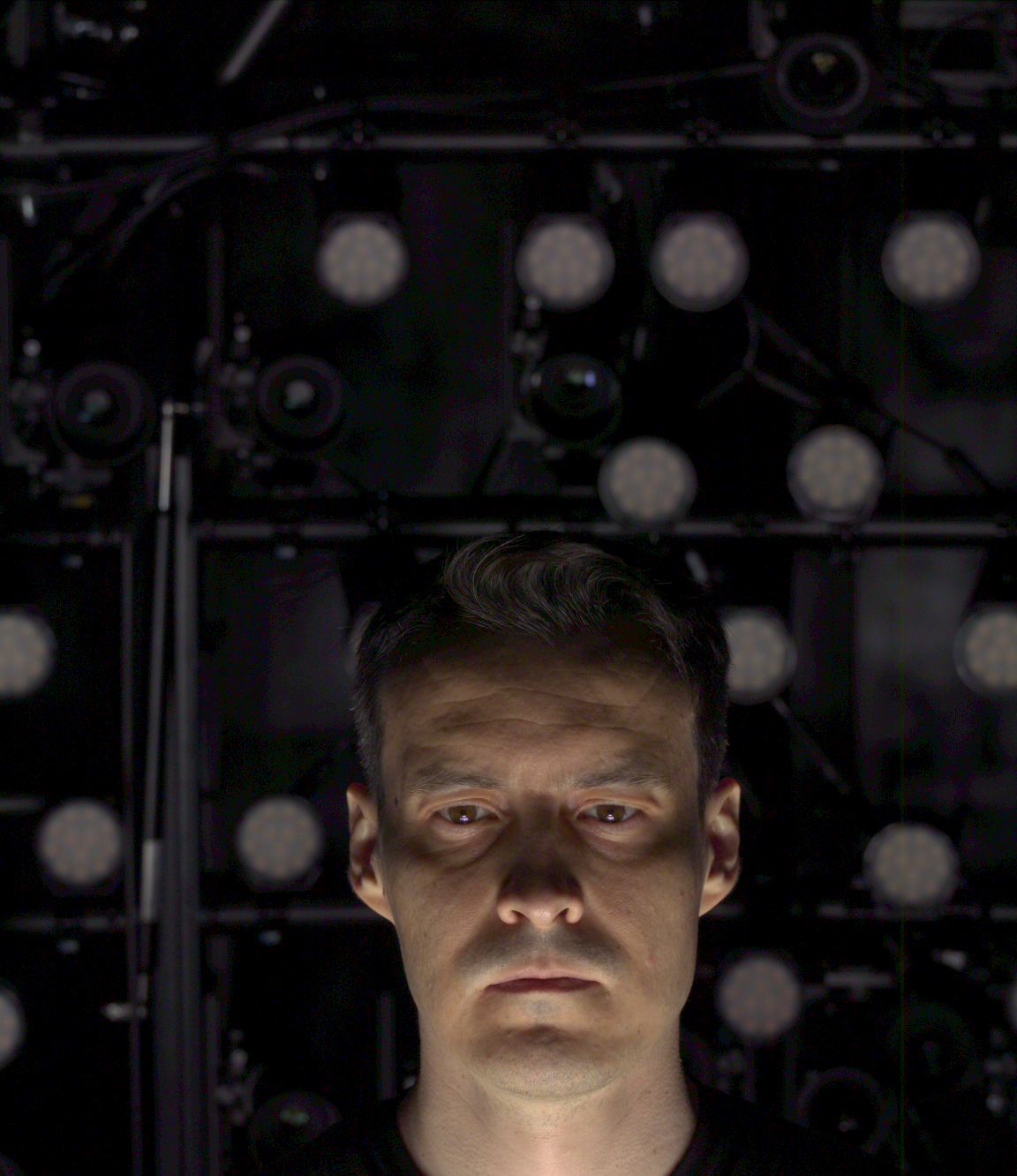}
        \caption{GT}
    \end{subfigure}
    \begin{subfigure}[b]{0.31\linewidth}
        \centering
        \includegraphics[width=\linewidth, trim=13cm 0cm 10cm 20cm, clip]{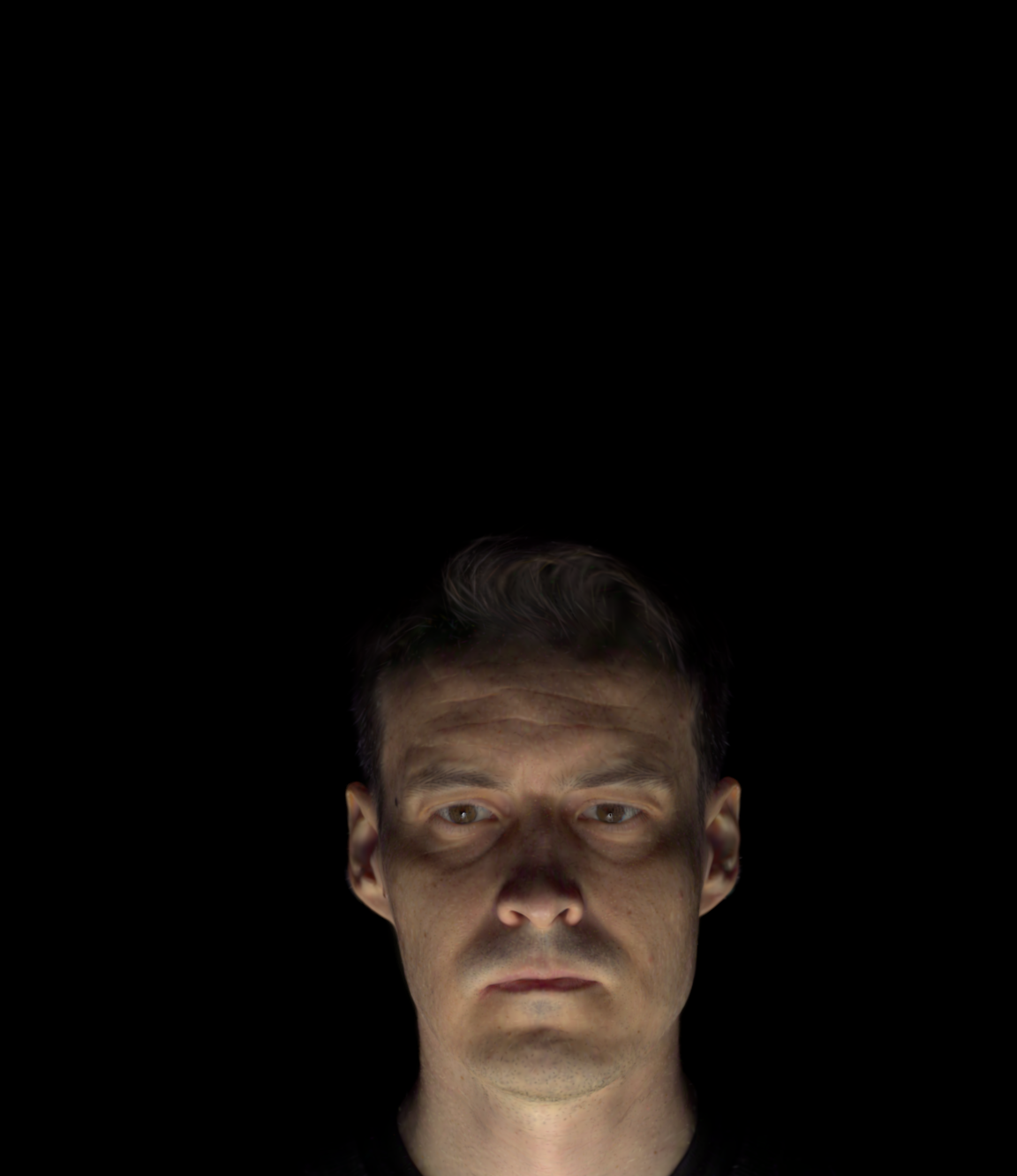}
        \caption{Ours (w. deferred)}
    \end{subfigure}
    \begin{subfigure}[b]{0.31\linewidth}
        \centering
        \includegraphics[width=\linewidth, trim=13cm 0cm 10cm 20cm, clip]{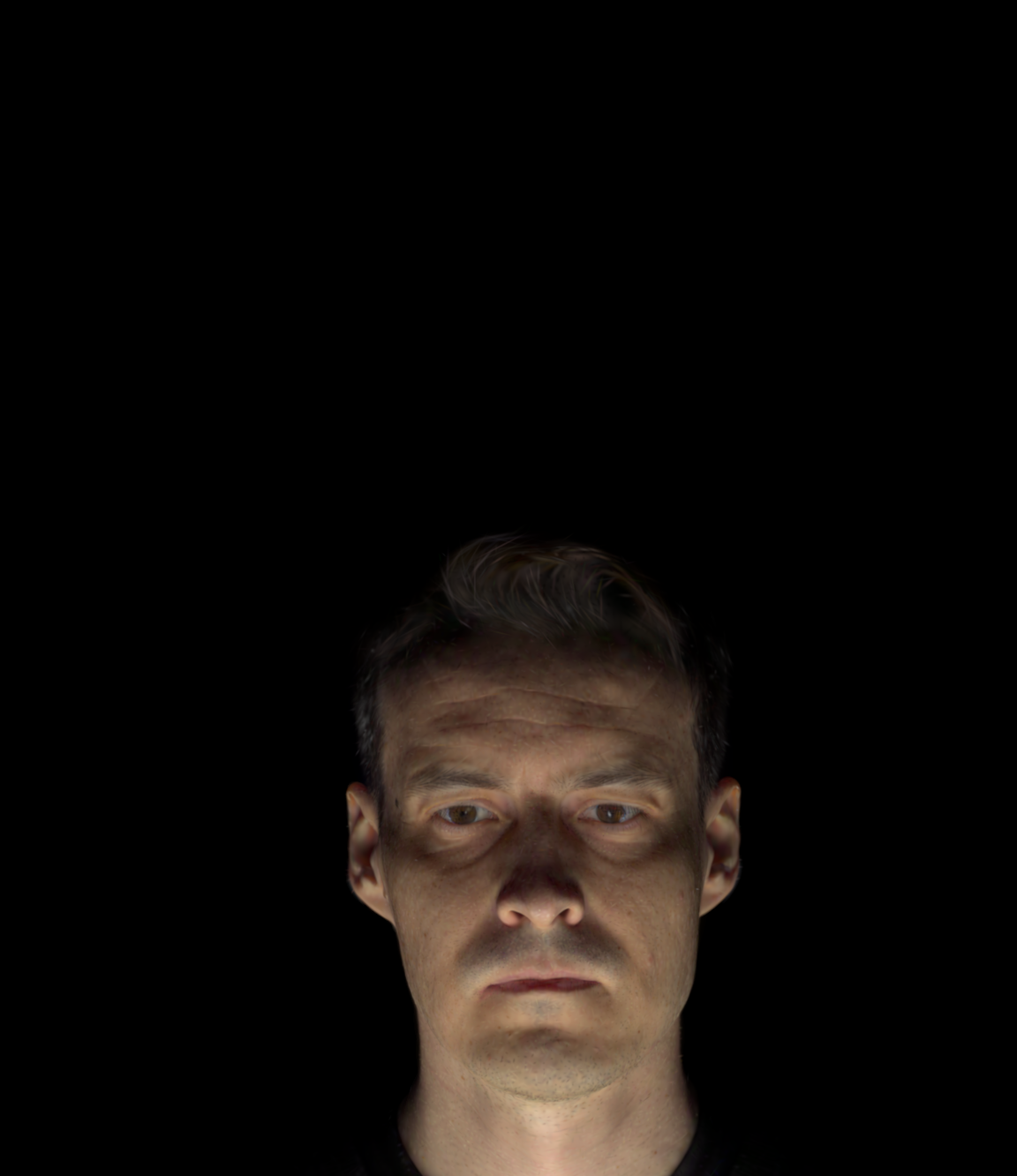}
        \caption{w.o. deferred}
    \end{subfigure}
    \caption{\textbf{Deferred shading.}  Without deferred shading, the specular
    reflections in eyes are either not captured or blurred.}
    \label{fig:qualitative_deferred}
\end{figure}

We note that the \textit{w.o. deferred} baseline achieves slightly better PSNR/SSIM compared to 
the full model.
This could be attributed to two reasons: 1) \textit{w.o. deferred} produces an overall smoother appearance due to alpha blending of multiple specular color predictions for a single pixel; metrics such as PSNR/SSIM often favor this kind of smoothed appearance, while LPIPS reflects more on the overall perceptual quality of the rendering.
This is demonstrated in Fig.~\ref{fig:qualitative_deferred} where \textit{w.o. deferred} misses high-frequency reflections on the nose and eyes.  2) Our current geometry formulation for deferred shading is
error-prone due to the noisy per-pixel depth estimation from Gaussian splatting.
Misalignment in depth could result in errors in surface-to-light vectors, and subsequently propagating to shading results.  The vanilla 3DGS is known for its under-representation of precise scene geometry.  Several recent works try to improve the geometry reconstruction of 3DGS~\cite{Huang2024SIGGRAPH,Yu2024SIGGRAPHASIA,Chen2024ARXIV}.
Incorporating these improvements in our geometry representation would be an interesting future work.

\section{Conclusion}
\label{sec:conclusion}
We have introduced a novel method for full-body, relightable, and drivable human
avatar reconstruction from light-stage data.  Our experiments show that
perceptually realistic relightable full-body avatars can be achieved by
combining a zonal-harmonic-based, orientation-dependent diffuse radiance
transfer, and a deferred-shading-based specular radiance transfer, all learned
from image observations only.  We have also demonstrated that non-local shadows
caused by body articulation can be captured by irradiance-conditioned shadow
networks.  Overall, our approach achieves a significant improvement in quality
for full-body relightable human avatar modeling, compared to existing PBR-based
models.

\boldparagraph{Limitations} Our method has several limitations.  First, the
cloth dynamics are based purely on the learned latent space, which may not be
physically plausible.  In such a case, the method may fail in
out-of-distribution scenarios, \eg\ when hands are touching the cloth or when
extreme body poses are present.  A more physically plausible clothing
layer~\cite{Xiang2023DrivableAC,Xiang2022DressingA,Rong2024ARXIV,Peng2024ARXIV,Zheng2024ECCV}
could be potentially integrated to resolve this issue.  Second, our method is
still suboptimal in capturing detailed appearances of eyes, faces, and hands
compared to specialized
methods~\cite{li2022eyenerf,Saito2024CVPR,Chen2024CVPR,Iwase2023CVPR}, as the
model capacity assigned to these regions is limited.  This could be potentially
solved by dynamically assigning UV space capacity to different body parts during
learning.  Lastly, our method has limited scalability as it requires a
multi-camera setup with known light sources, a natural future direction is to
extend the method to universal setups similar to related
face~\cite{Li2024SIGGRAPHASIA} and hand~\cite{Chen2024CVPR} models.

\bibliographystyle{ACM-Reference-Format}
\bibliography{ref}

\clearpage
\section*{}
\begin{figure*}
    \centering
    \begin{subfigure}[b]{0.24\linewidth}
        \centering
        \includegraphics[width=\linewidth, trim=0cm 0cm 0cm 0cm, clip]{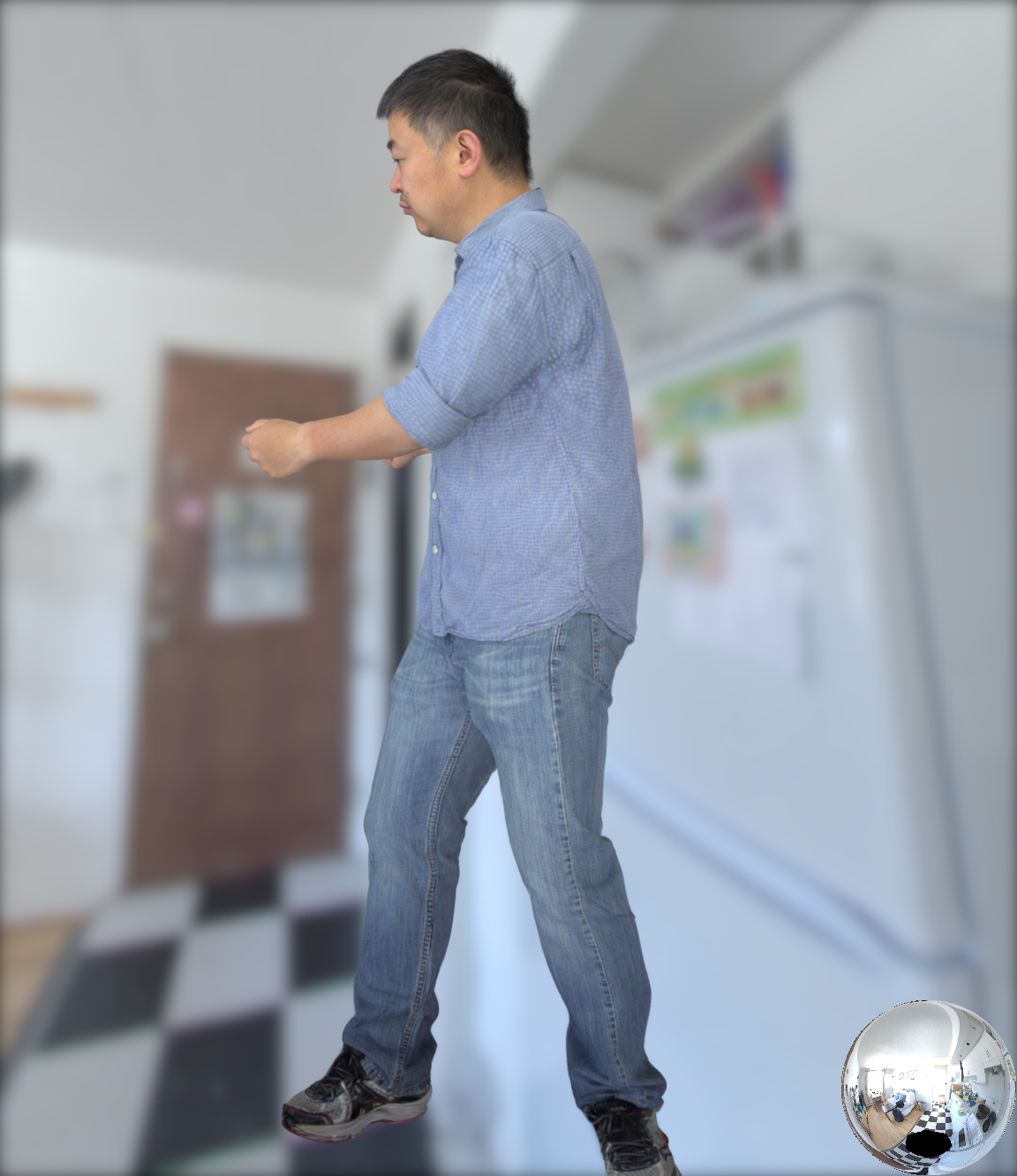}
    \end{subfigure}
    \begin{subfigure}[b]{0.24\linewidth}
        \centering
        \includegraphics[width=\linewidth]{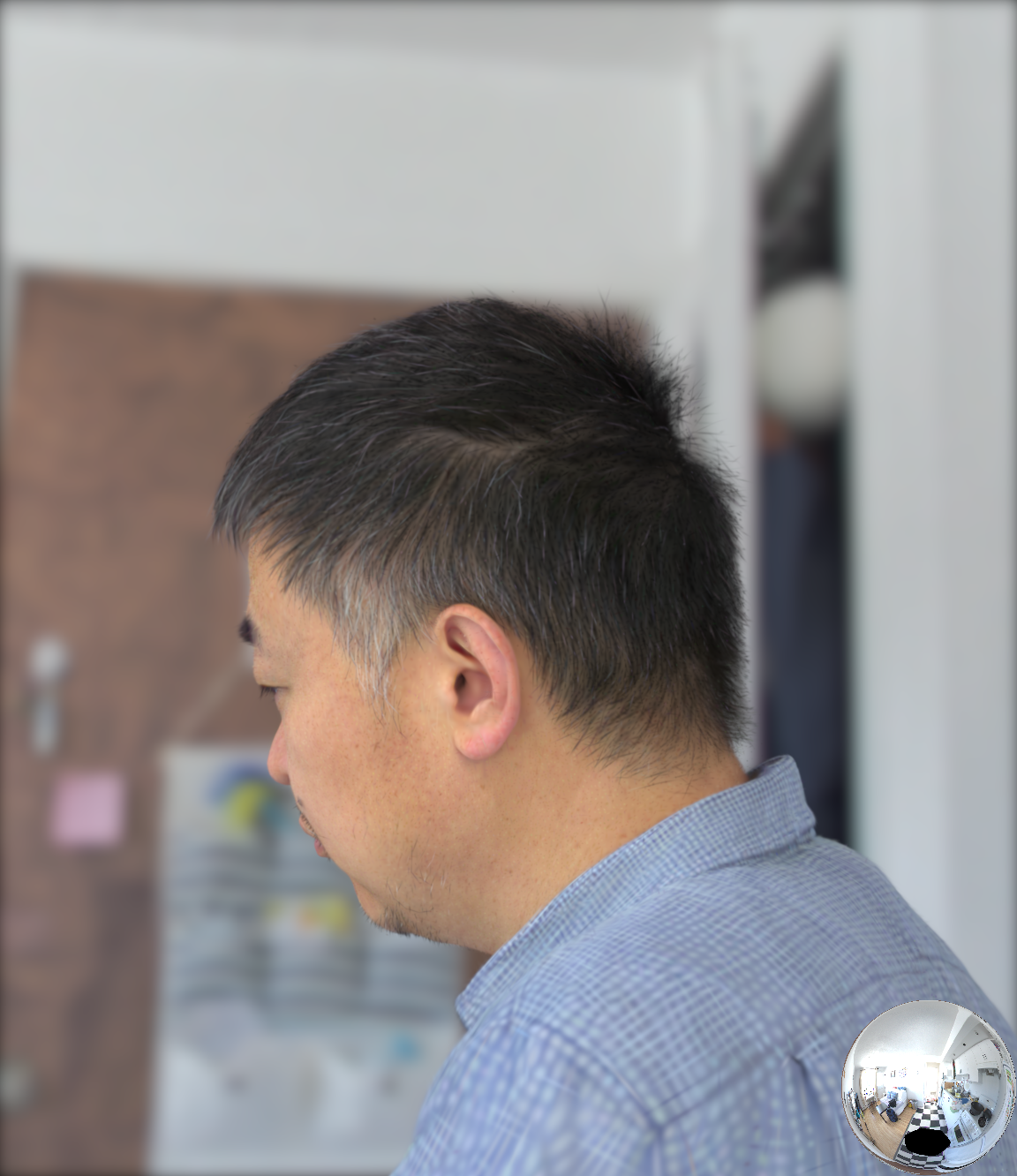}
    \end{subfigure}
    \begin{subfigure}[b]{0.24\linewidth}
        \centering
        \includegraphics[width=\linewidth]{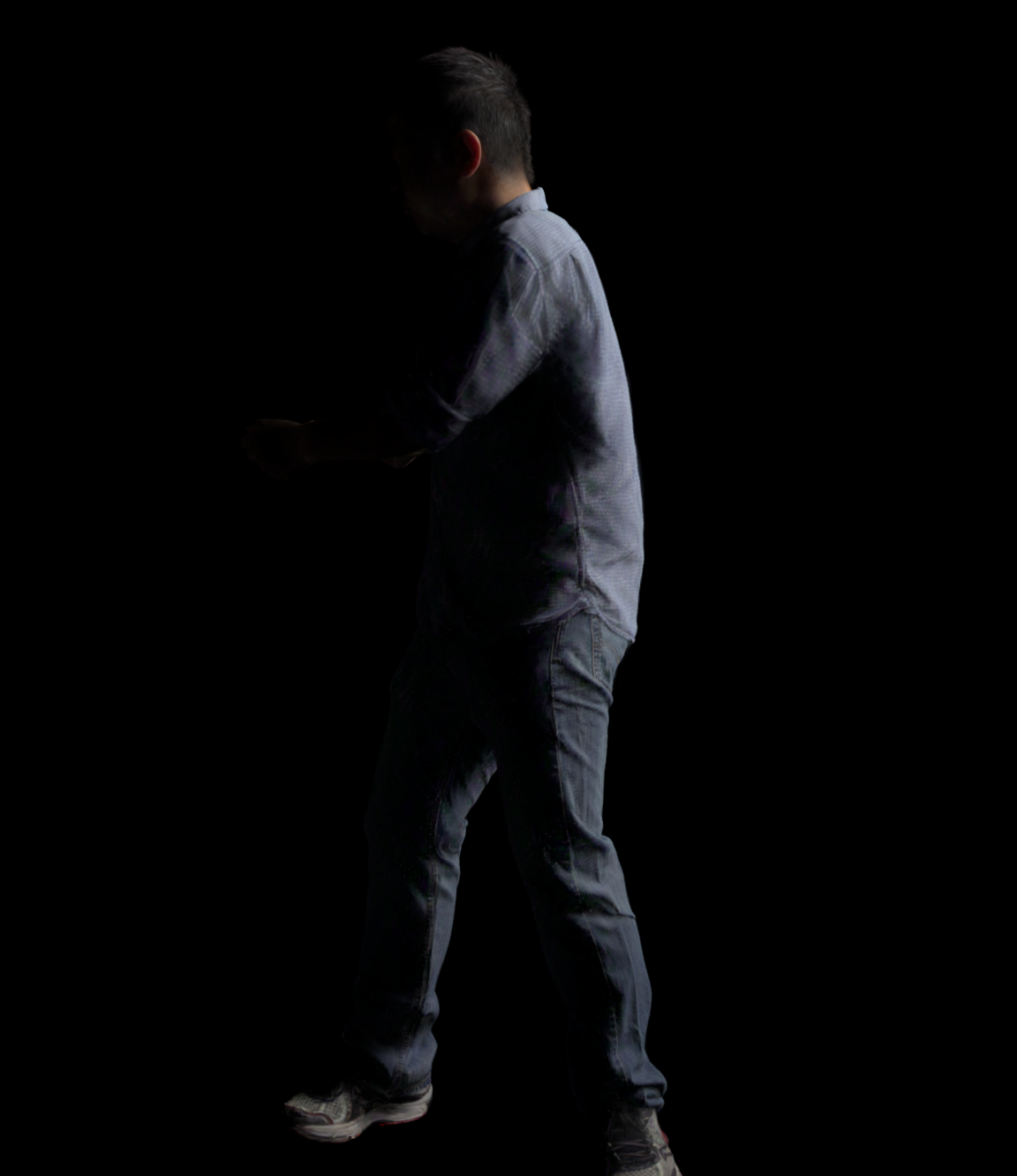}
    \end{subfigure}
    \begin{subfigure}[b]{0.24\linewidth}
        \centering
        \includegraphics[width=\linewidth]{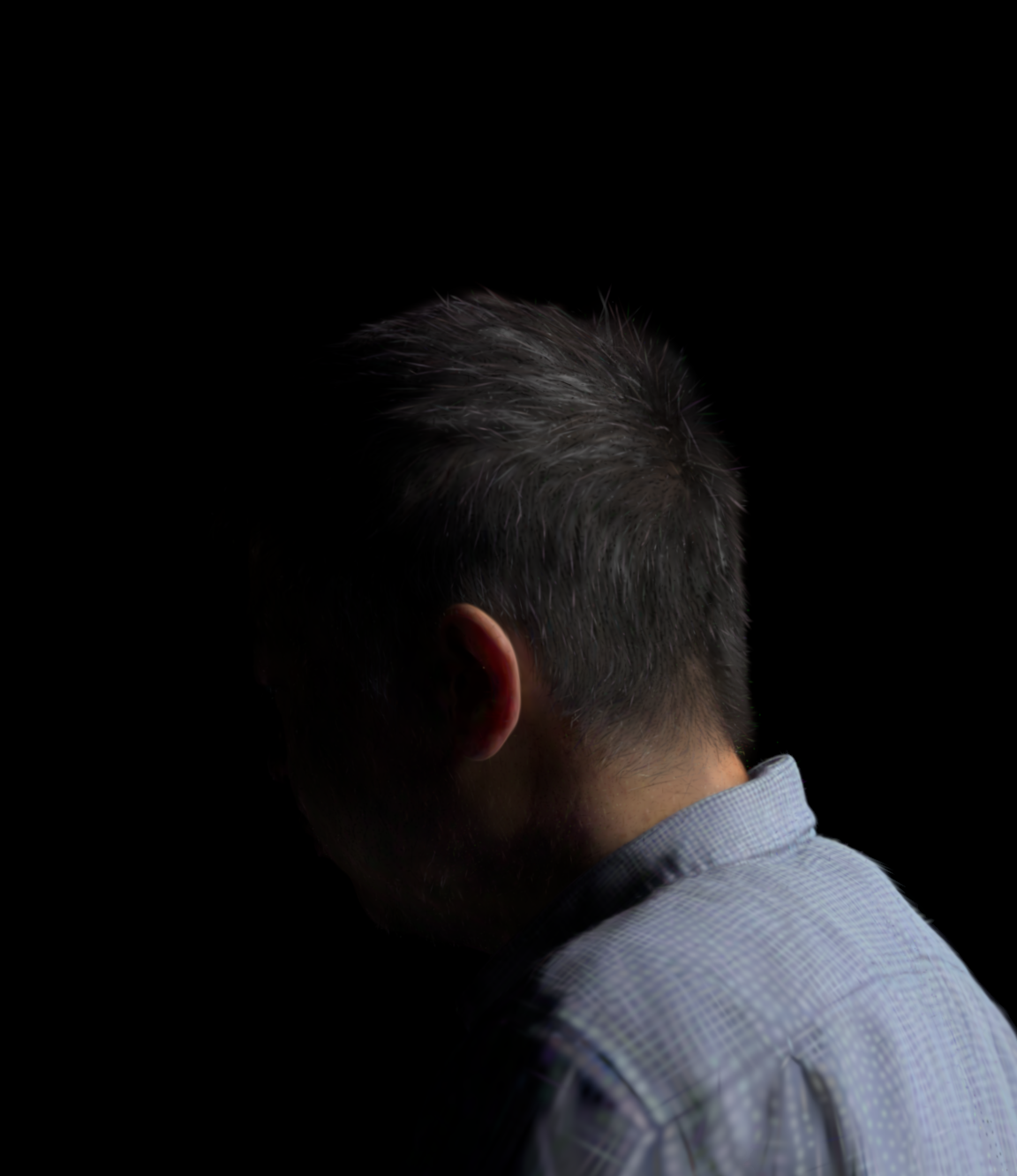}
    \end{subfigure}
    \begin{subfigure}[b]{0.24\linewidth}
        \centering
        \includegraphics[width=\linewidth]{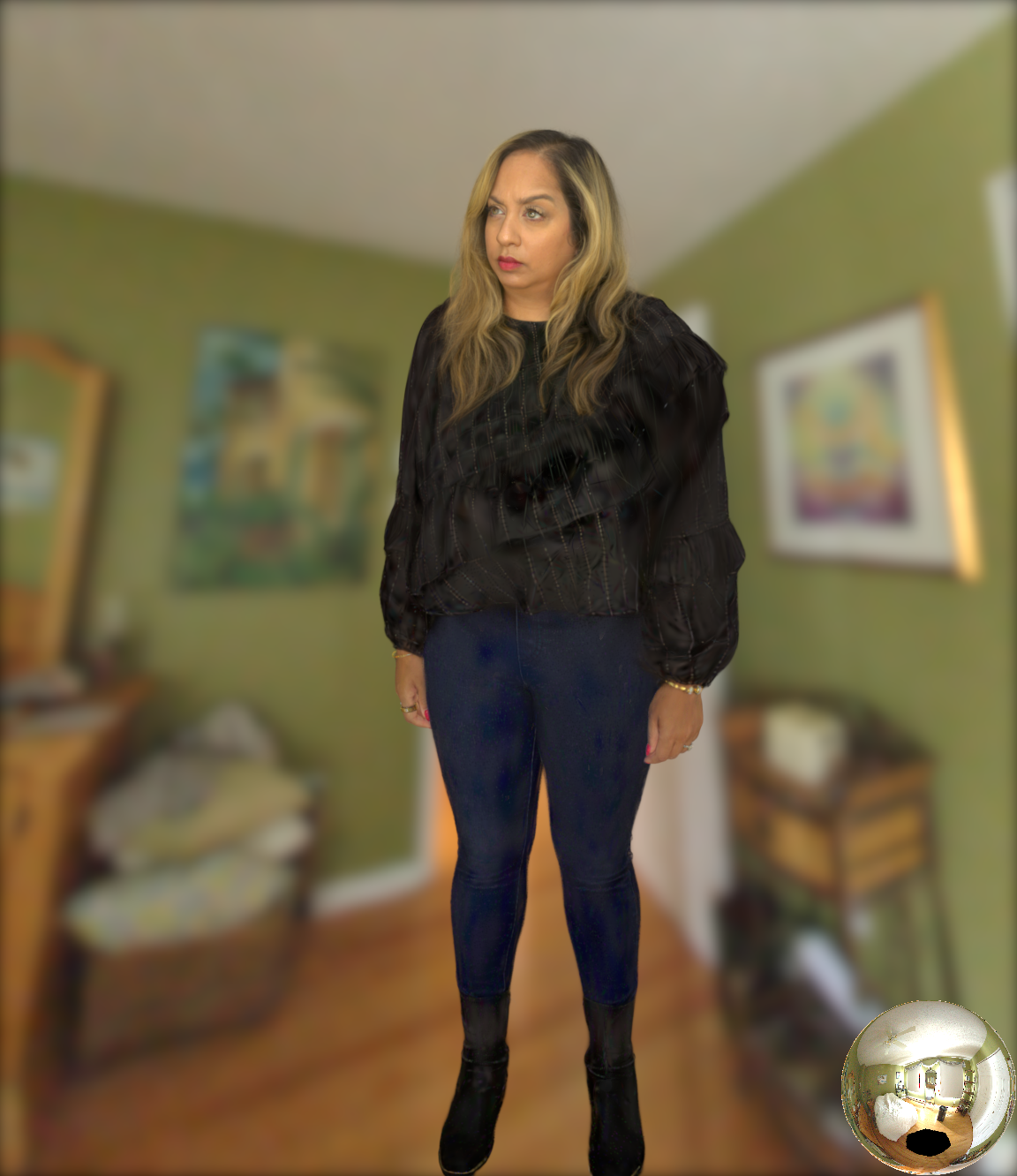}
    \end{subfigure}
    \begin{subfigure}[b]{0.24\linewidth}
        \centering
        \includegraphics[width=\linewidth]{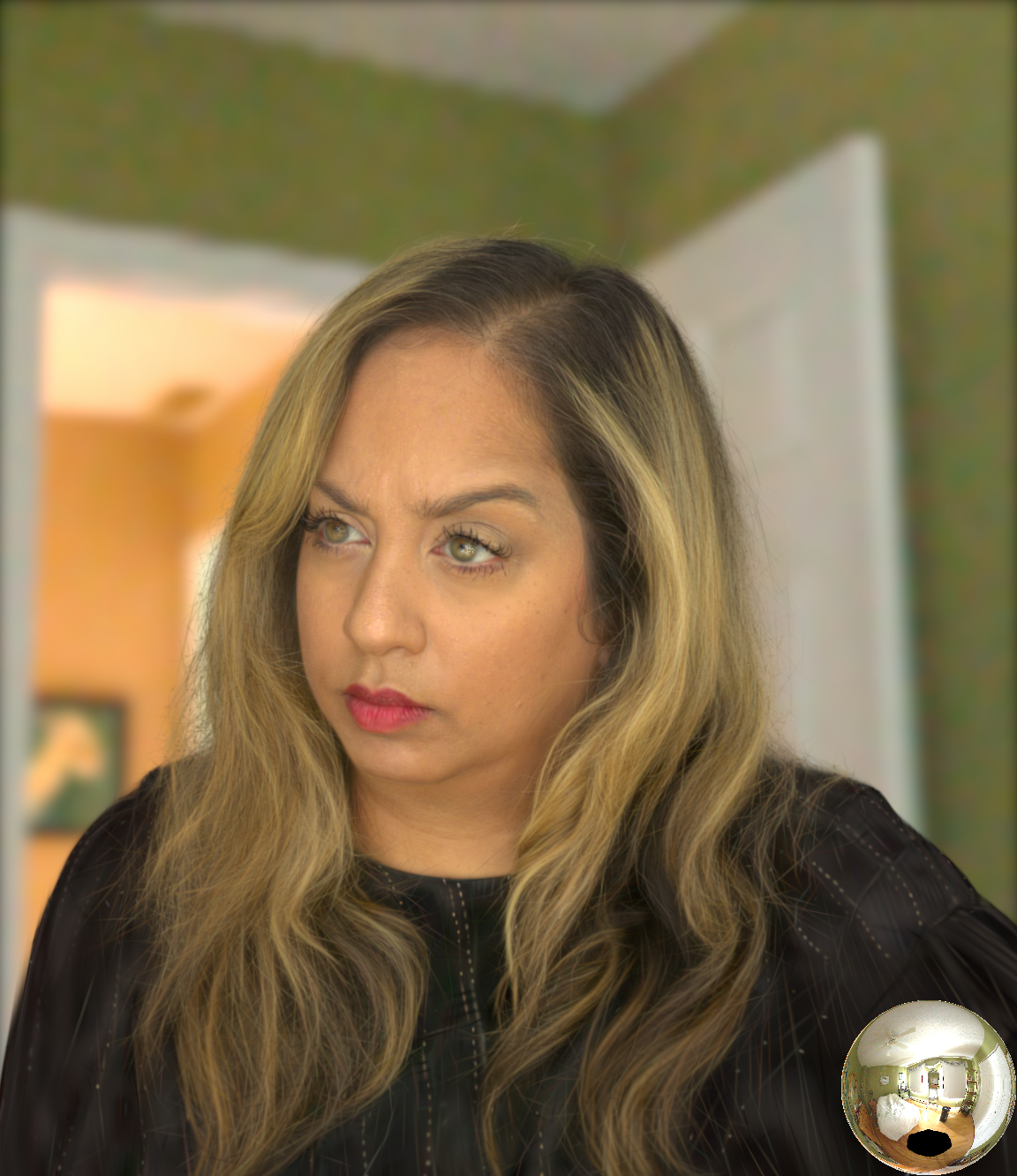}
    \end{subfigure}
    \begin{subfigure}[b]{0.24\linewidth}
        \centering
        \includegraphics[width=\linewidth]{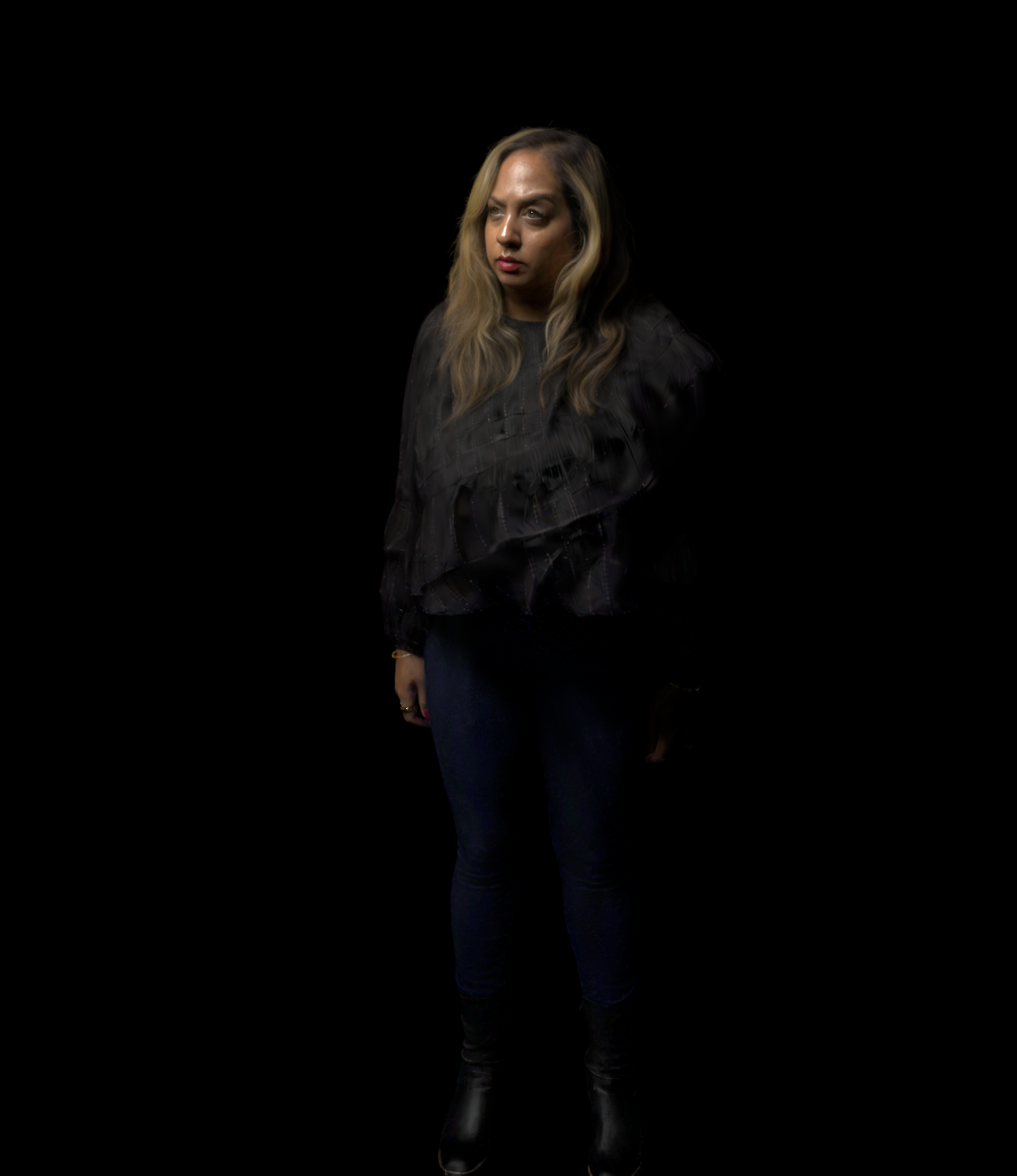}
    \end{subfigure}
    \begin{subfigure}[b]{0.24\linewidth}
        \centering
        \includegraphics[width=\linewidth]{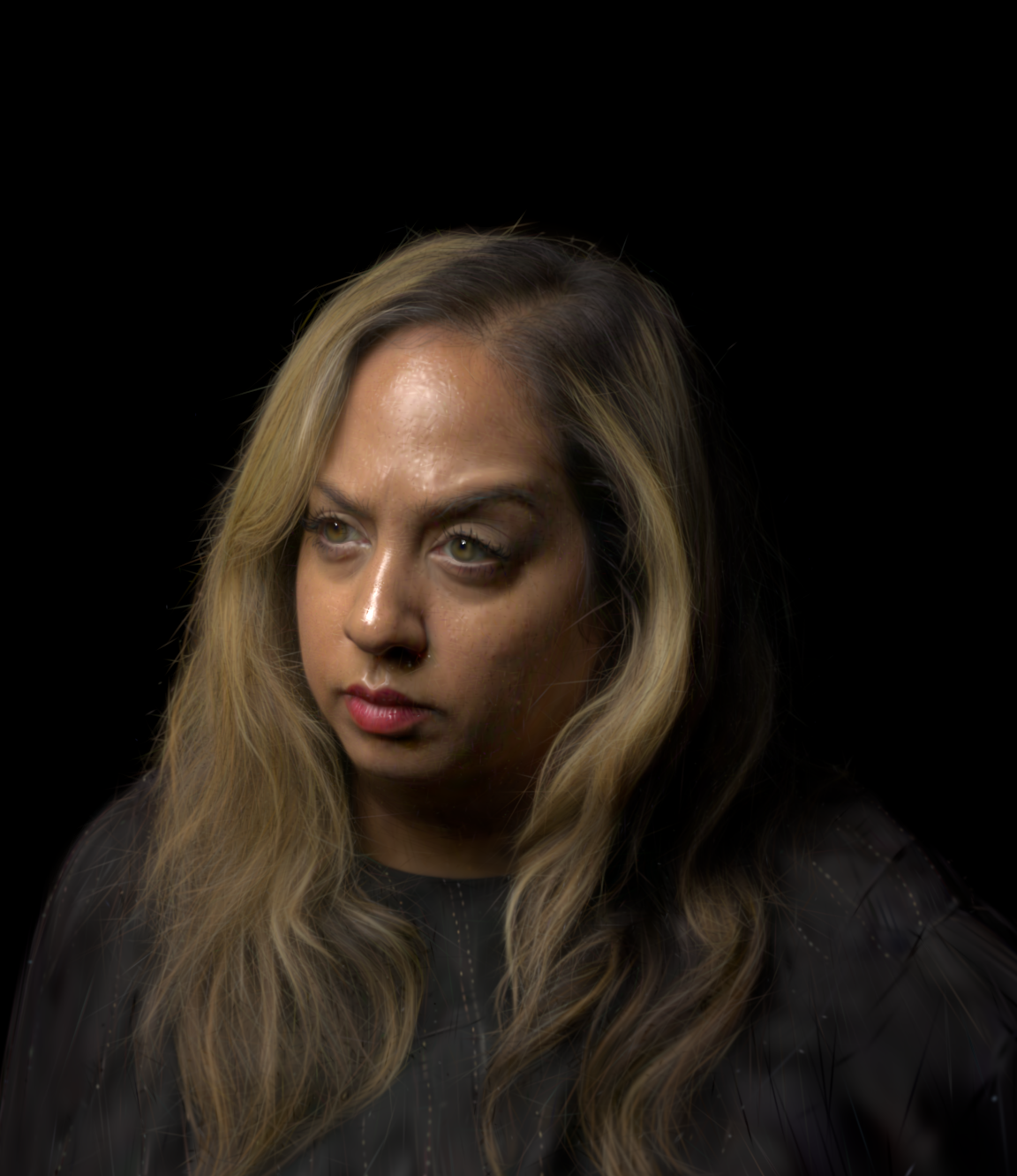}
    \end{subfigure}
    \begin{subfigure}[b]{0.24\linewidth}
        \centering
        \includegraphics[width=\linewidth]{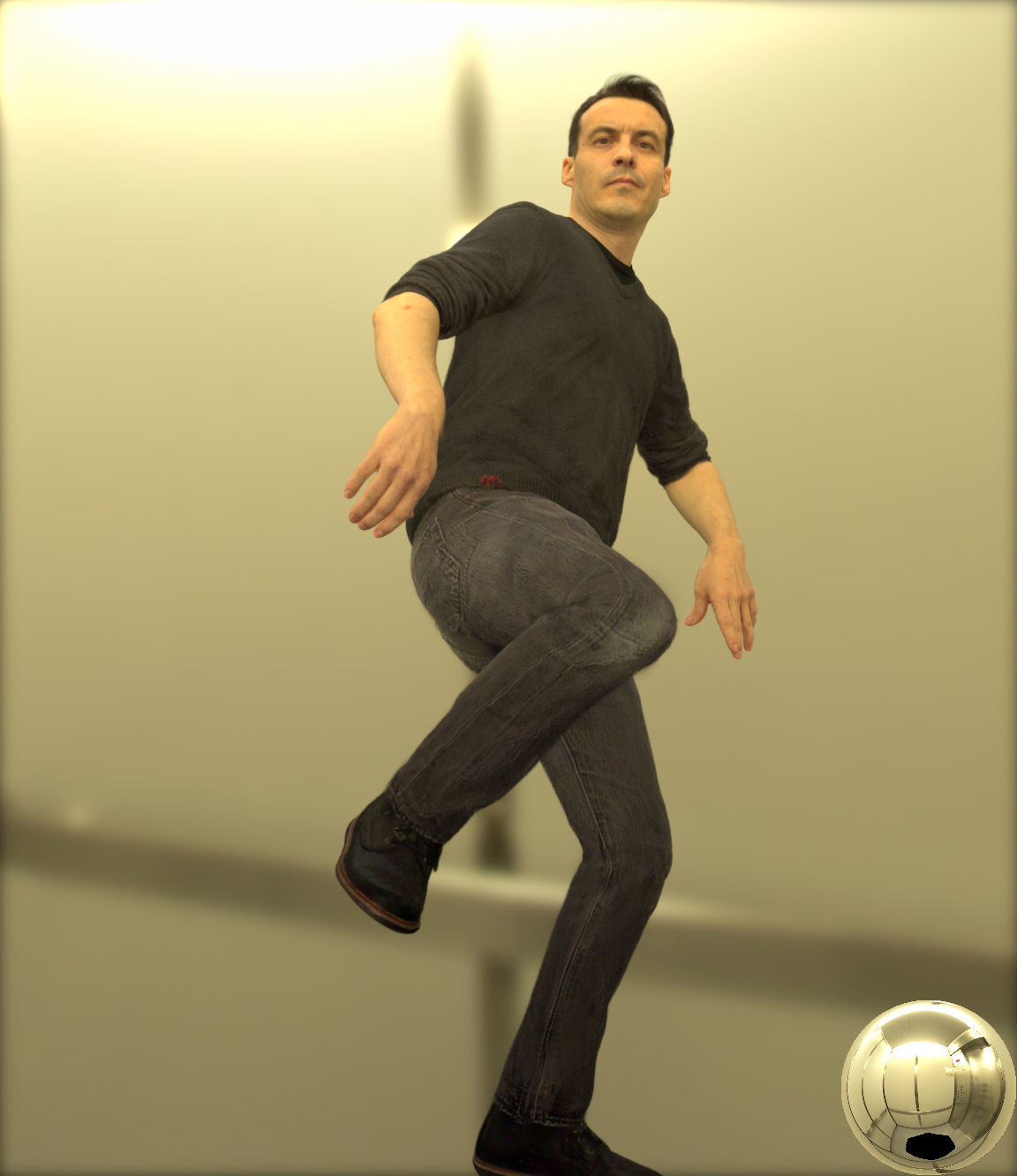}
    \end{subfigure}
    \begin{subfigure}[b]{0.24\linewidth}
        \centering
        \includegraphics[width=\linewidth]{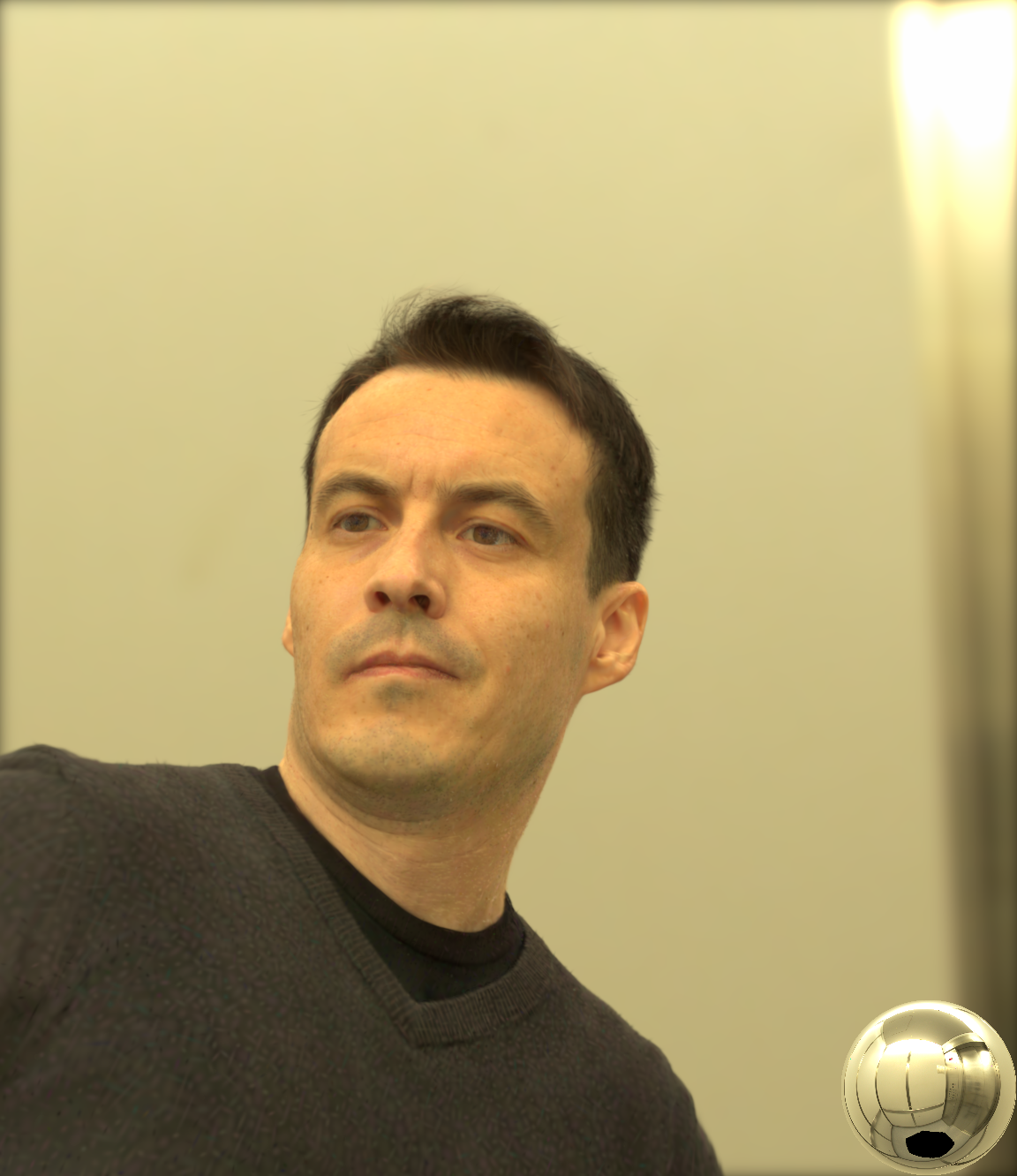}
    \end{subfigure}
    \begin{subfigure}[b]{0.24\linewidth}
        \centering
        \includegraphics[width=\linewidth]{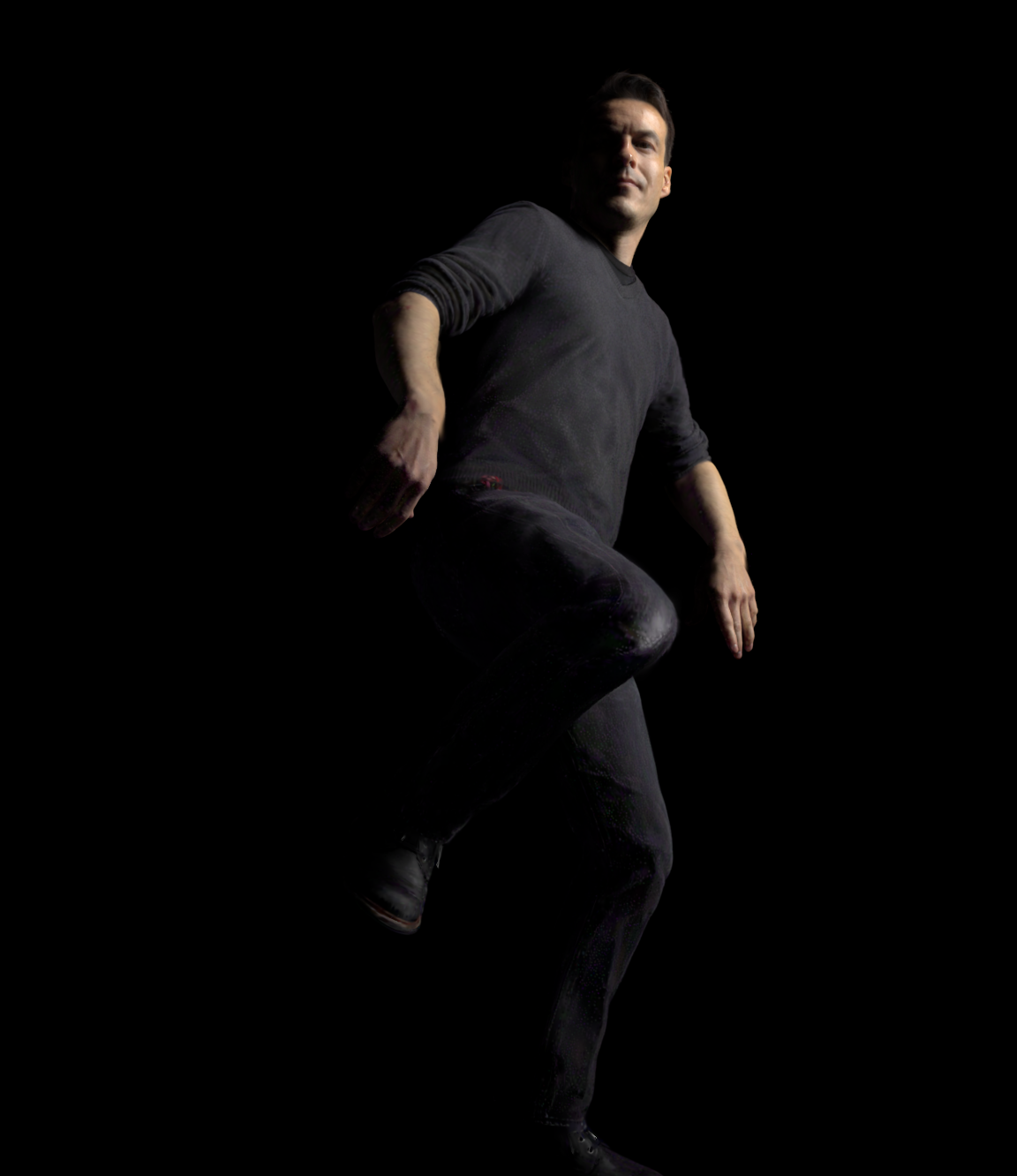}
    \end{subfigure}
    \begin{subfigure}[b]{0.24\linewidth}
        \centering
        \includegraphics[width=\linewidth]{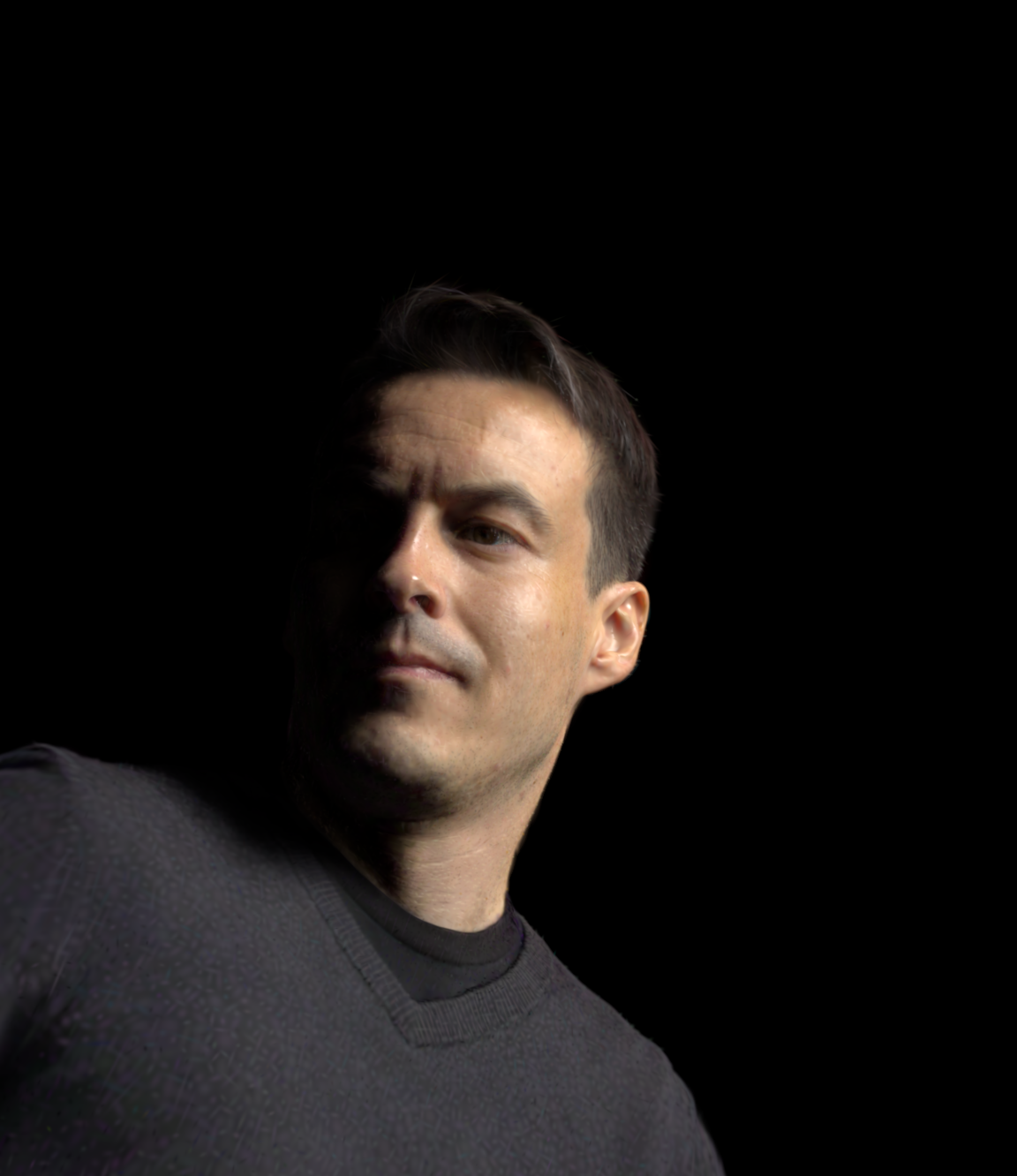}
    \end{subfigure}
    \begin{subfigure}[b]{0.24\linewidth}
        \centering
        \includegraphics[width=\linewidth]{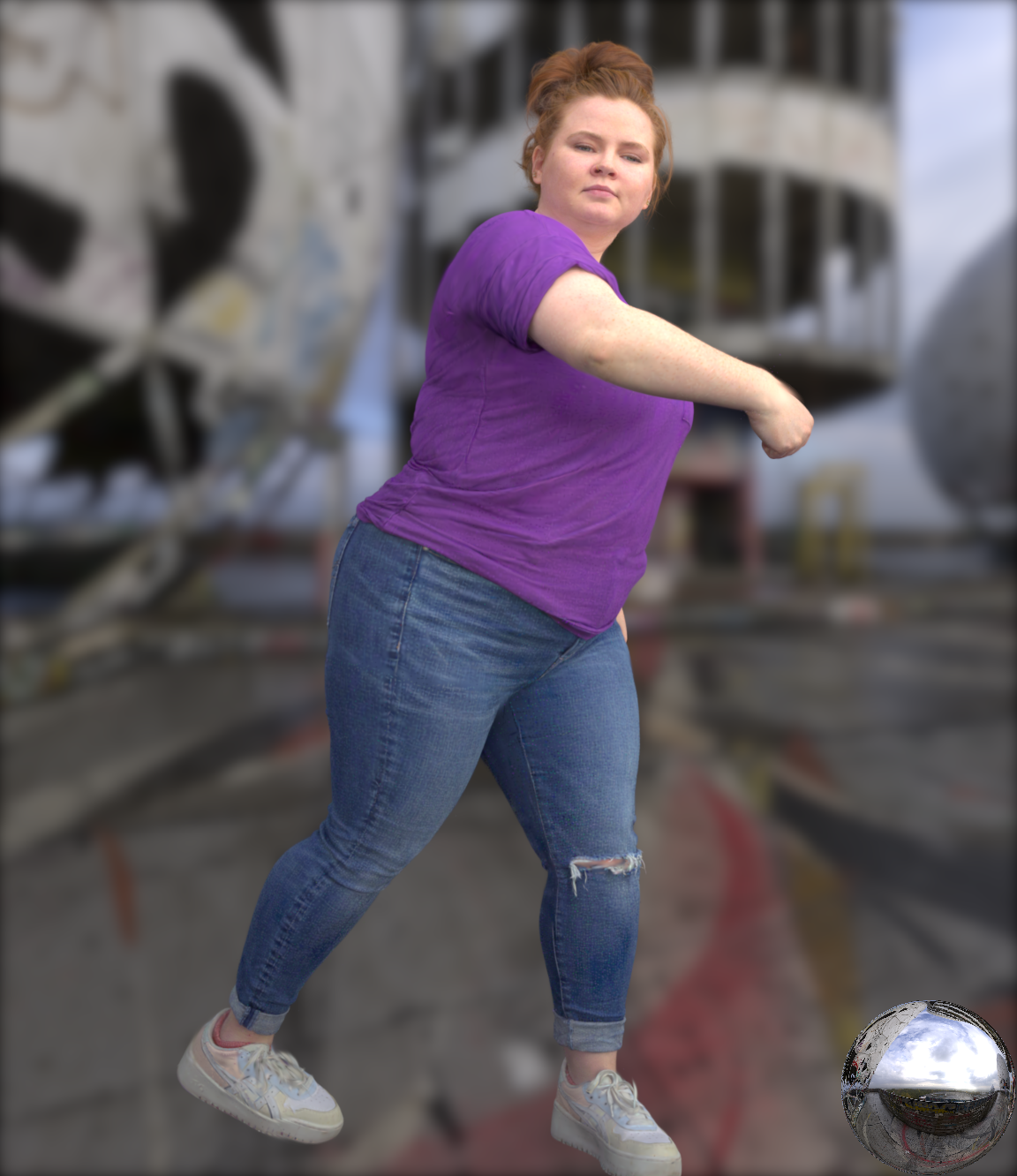}
    \end{subfigure}
    \begin{subfigure}[b]{0.24\linewidth}
        \centering
        \includegraphics[width=\linewidth]{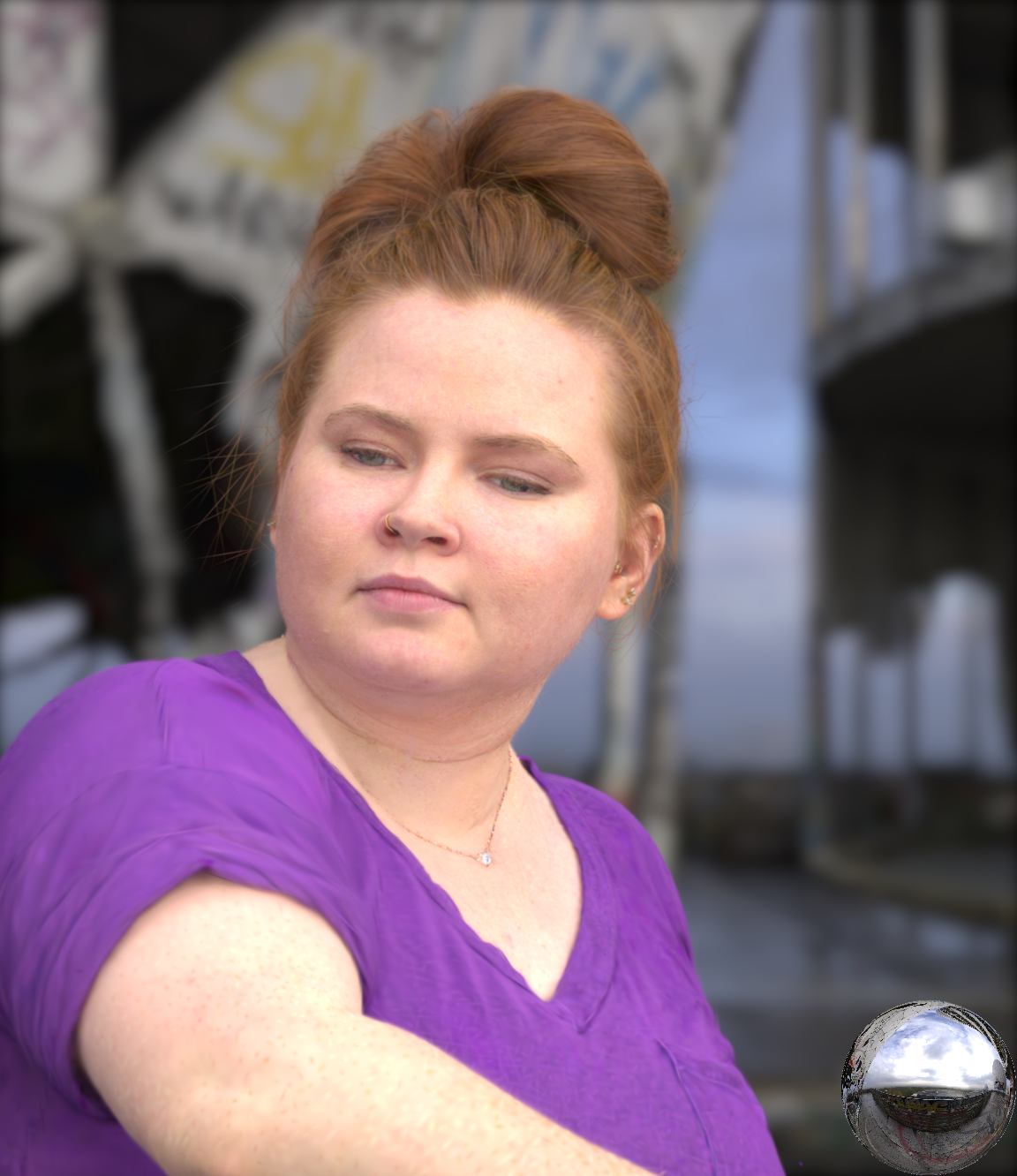}
    \end{subfigure}
    \begin{subfigure}[b]{0.24\linewidth}
        \centering
        \includegraphics[width=\linewidth]{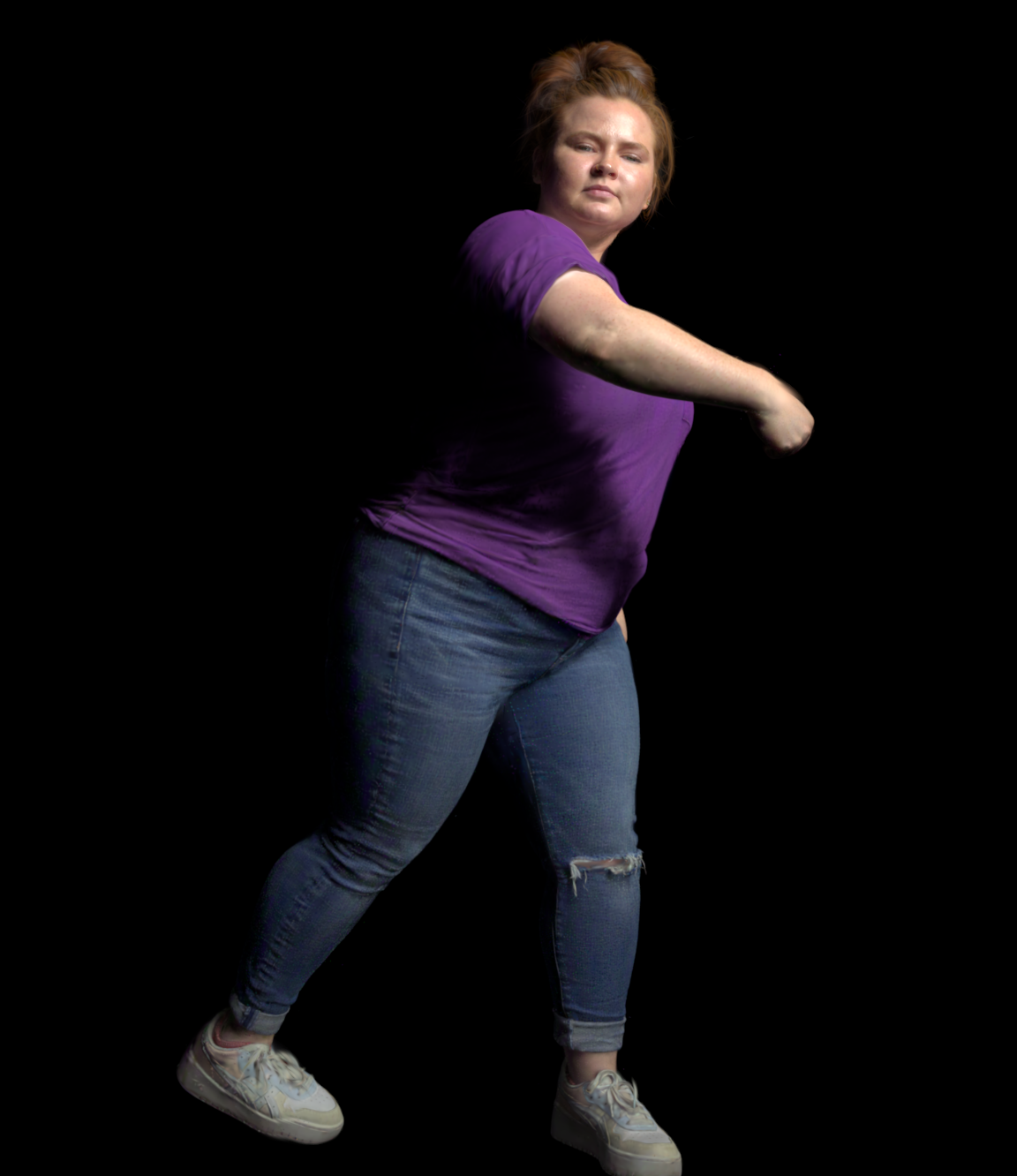}
    \end{subfigure}
    \begin{subfigure}[b]{0.24\linewidth}
        \centering
        \includegraphics[width=\linewidth]{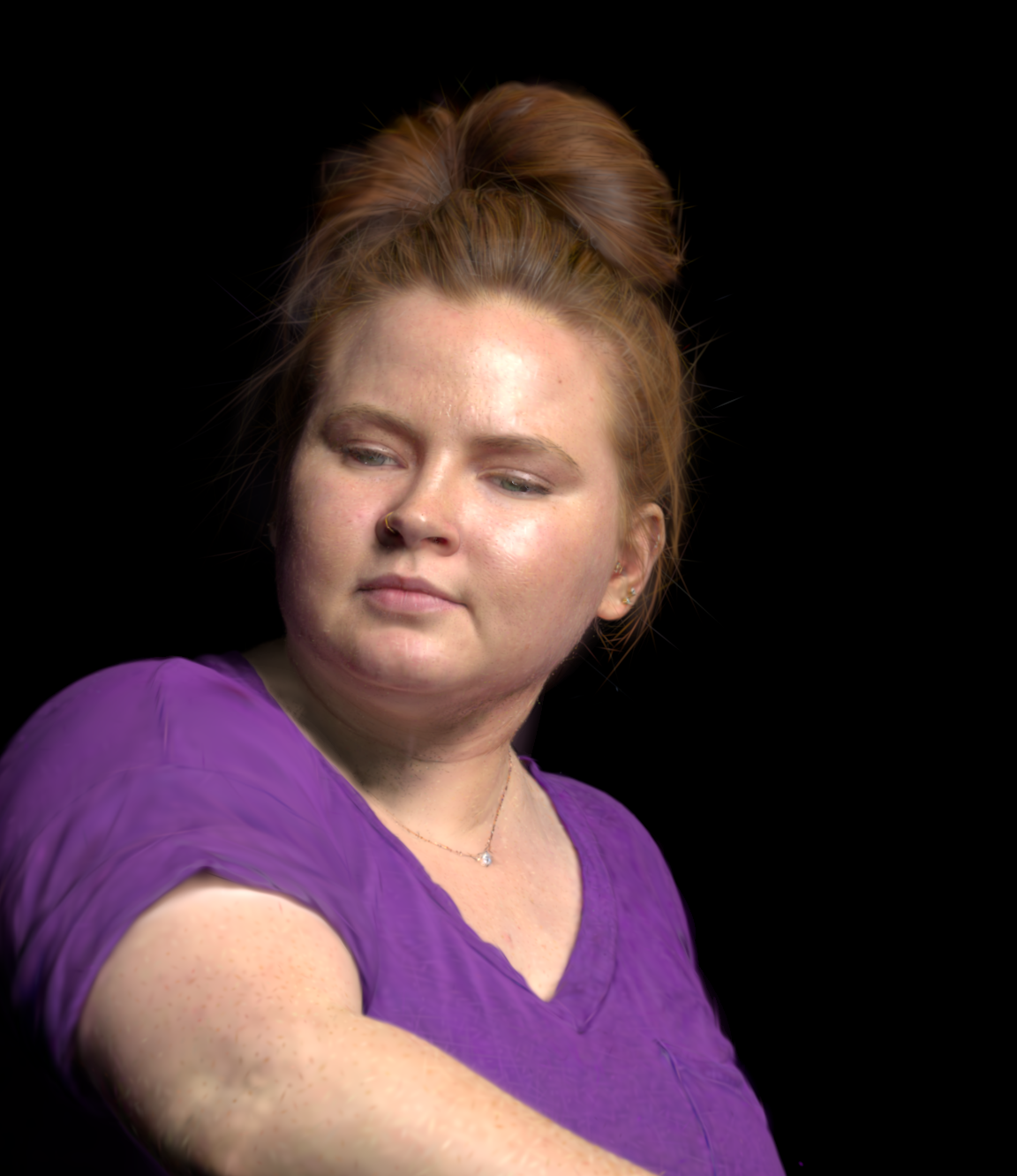}
    \end{subfigure}
    \caption{\textbf{Relighting result on unseen motion.}  We show environment-map-based relighting on the left two columns and point-light-based relighting on the right two columns.}
    \label{fig:qualitative_relight}
\end{figure*}

\appendix
\onecolumn
\numberwithin{equation}{section}
\setcounter{equation}{0}
\numberwithin{figure}{section}
\setcounter{figure}{0}
\numberwithin{table}{section}
\setcounter{table}{0}

\section{Monte Carlo Integration for Normalized Irradiance}
\label{appx:mc_integration}
In Sec.~\iftoggle{arXiv}{\ref{sec:shadowing}}{3.3 of the main paper}, we
proposed to compute the normalized irradiance as follows:
\begin{align}
    \label{appx:eqn:shadowing}
    \text{Irradiance} (\mathbf{v}_k) = \frac{\int_{\mathbb{S}^2} \mathbf{L}(\mathbf{v}_k, \mathbf{\omega}_i) \text{Vis} (\mathbf{v}_k, \mathbf{\omega}_i) \text{d} \mathbf{\omega}_i}{\int_{\mathbb{S}^2} \mathbf{L}(\mathbf{v}_k, \mathbf{\omega}_i) \text{d} \mathbf{\omega}_i}
\end{align}
in practice, assume we have in total $M$ light sources (either the number of point lights, or
the number of pixels on an environment map), such normalized irradiance can be approximated with
Monte Carlo integration:
\begin{align}
    \label{appx:eqn:shadowing_mc}
    \text{Irradiance} (\mathbf{v}_k) &\approx \frac{ \sum_{j=1}^{N} \frac{1}{N} \frac{\mathbf{L}(\mathbf{v}_k, \mathbf{\omega}_i^j) \text{Vis} (\mathbf{v}_k, \mathbf{\omega}_i^j)}{\text{pdf} (\mathbf{\omega}_i^j)}}{ \sum_{j=1}^{M} \frac{1}{M} \frac{ \mathbf{L}(\mathbf{v}_k, \mathbf{\omega}_i^j)}{\bar{\text{pdf}} (\mathbf{\omega}_i^j)}}
\end{align}
where $\{ \mathbf{\omega}_i^j \}_{j=1}^{N}$ are $N$ samples drawn via light
importance sampling, $\{ \text{pdf} (\mathbf{\omega}_i^j) \}_{j=1}^{N}$ are the
corresponding PDF values.  $\{ \mathbf{\omega}_i^j \}_{j=1}^{M}$ are directions
towards each of the light sources, $\{ \bar{\text{pdf}} (\mathbf{\omega}_i^j) \}_{j=1}^{M}$
are the corresponding PDF values.  The denominator can always computed efficiently
as it does not include the visibility term.
For light-stage data, we have $M$ light sources with equal
light intensity, thus $\bar{\text{pdf}} (\cdot) \equiv \text{pdf} (\cdot) \equiv \frac{1}{M}$.  We can further
simplify the above equation to:
\begin{align}
    \label{appx:eqn:shadowing_light_stage}
    \text{Irradiance} (\mathbf{v}_k) &\approx \frac{ \sum_{j=1}^{N} \frac{M}{N} \mathbf{L}(\mathbf{v}_k, \mathbf{\omega}_i^j) \text{Vis} (\mathbf{v}_k, \mathbf{\omega}_i^j)}{ \sum_{j=1}^{M} \mathbf{L}(\mathbf{v}_k, \mathbf{\omega}_i^j)}
\end{align}
We also note that Eq.~\eqref{appx:eqn:shadowing_mc} can be computed with a reduced
number of samples per pixel $N$.  Reducing $N$ will not change the
expectation of the result but will increase the variance.  On the other hand,
the normalized irradiance maps are inputs to the neural network, which could
potentially serve as a denoiser.  Indeed, we demonstrate in
Table~\ref{appx:tab:quantitative} that even using 1 sample per pixel (1SPP) for
approximating Eq.~\eqref{appx:eqn:shadowing}, the accuracy drop is minimal while the
computational cost is significantly reduced.
\begin{table}
\small
\renewcommand{\tabcolsep}{2.0pt}
    \centering
    \begin{tabular}{l|c|c|c|c|c|c|}
        \toprule
        \multirow{2}{*}{Method} & \multicolumn{3}{c|}{Training Motion} & \multicolumn{3}{c}{Unseen Motion} \\
        \cmidrule{2-7}
        & PSNR $\uparrow$ & SSIM $\uparrow$ & LPIPS $\downarrow$ & PSNR $\uparrow$ & SSIM $\uparrow$ & LPIPS $\downarrow$ \\
        \midrule
        PBR            & 28.35 & 0.7729 & 0.1993 & 26.83 & 0.7477 & 0.2166 \\
        \midrule
        SH             & 29.15 & 0.7958 & 0.1846 & 27.21 & 0.7679 & 0.2056 \\
        w.o. shadow    & 28.89 & 0.7991 & 0.1800 & 27.07 & 0.7707 & 0.2004 \\
        w.o. deferred  & \cellcolor{rred} 29.55 & \cellcolor{rred} 0.8047 & 0.1796 & \cellcolor{rred} 27.59 & \cellcolor{oorange} 0.7755 & 0.2003 \\
        Mesh normal    & 29.43 & 0.8036 & 0.1785 & 27.53 & 0.7747 & 0.1993 \\
        \midrule
        Ours           & \cellcolor{oorange} 29.48 & \cellcolor{oorange} 0.8046 & \cellcolor{rred} 0.1781 & \cellcolor{oorange} 27.57 &\cellcolor{rred}  0.7756 & \cellcolor{oorange} 0.1989 \\
        Ours (1spp)    & 29.40 & 0.8031 & \cellcolor{oorange} 0.1782 & 27.53 & 0.7744 & \cellcolor{rred} 0.1988 \\
        \bottomrule
    \end{tabular}
    \caption{\textbf{Quantitative comparison to baselines.}  The top two 
 approaches are highlighted in \textcolor{rred}{red} and \textcolor{oorange}{orange}, respectively.}
    \label{appx:tab:quantitative}
\end{table}

\section{Loss Definition}
\label{appx:loss_definition}
In this section, we extend Sec.~\iftoggle{arXiv}{\ref{sec:losses}}{3.4 of the main paper} to include
detailed definitions of the regularization losses used in our method.  The regularization losses, as
defined in Eq.~\iftoggle{arXiv}{\eqref{eqn:reg_loss}}{18 of the main paper}, are as follows:
\begin{align}
    \mathcal{L}_{\text{reg}} =& \mathcal{L}_{\text{scale}} + \lambda_{\text{offset}} \mathcal{L}_{\text{offset}} + \lambda_{\text{mask}} \mathcal{L}_{\text{mask}} + \lambda_{\text{normal}} \mathcal{L}_{\text{normal}} \nonumber \\
    &+ \lambda_{\text{bound}} \mathcal{L}_{\text{bound}} + \lambda_{\text{normal\_orient}} \mathcal{L}_{\text{normal\_orient}} \nonumber \\
    &+ \lambda_{\text{alpha\_sparsity}} \mathcal{L}_{\text{alpha\_sparsity}} + \lambda_{\text{albedo}} \mathcal{L}_{\text{albedo}} \nonumber \\
    &+ \lambda_{\text{neg\_color}} \mathcal{L}_{\text{neg\_color}}
\end{align}
where $\mathcal{L}_{\text{scale}}$ is the L1 loss on the scale of the Gaussians
$\{ \mathbf{s}_k \}$.  $\mathcal{L}_{\text{offset}}$ is the L2 loss on the
predicted delta translations $\{ \delta \mathbf{t}_k \}$.
$\mathcal{L}_{\text{mask}}$ is the L1 mask loss between the rendered alpha maks
from Gaussian primitives and the ground truth segmentation mask.  Note that to
keep fine-scale details such as hairs, we exclude boundary regions of the
segmentation mask from the mask loss.  $\mathcal{L}_{\text{normal}}$ is the L2
loss on the predicted specular normal offsets $\{ \delta \mathbf{n}_k \}$.
$\mathcal{L}_{\text{bound}}$ penalizes Gaussian scales and roughness values that
go beyond predefined bounds.
Specifically:
\begin{align}
    \mathcal{L}_{\text{bound}} = \text{mean}( l_{\text{bound}} ), l_{\text{bound}} =
    \begin{cases}
        1 / \max (v, 10^{-7}) & \text{if} \quad v < lb \\
        (v - ub)^2 & \text{if} \quad v > ub
    \end{cases}
\end{align}
where $v$ are either Gaussian scales for each rotation axis or roughness
values.  $lb$ and $ub$ are the lower and upper bounds, respectively.  We set
$lb = 0.0001, ub =0.01$ for scales and $lb = 0.01, ub = 0.25$ for roughness.

$\mathcal{L}_{\text{normal\_orient}}$ is the squared loss on the dot product
between the deferred specular normals
(Eq.~\iftoggle{arXiv}{\eqref{eqn:deferred_normals}}{11 of the main paper}) and
the view directions $\mathbf{\omega}_o (u,v)$:
\begin{align}
    \mathcal{L}_{\text{normal\_orient}} = \text{mean} \left( \max (0, \hat{\mathbf{N}} (u, v) \cdot \mathbf{\omega}_o) (u,v) \right)^2
\end{align}
Further, $\mathcal{L}_{\text{alpha\_sparsity}}$ is the L1 loss on the alpha mask
to encourage alpha values to be either 0 or 1.  Note that we only apply this
loss for non-hair regions, as hair regions are expected to have non-binary
opacity values.  $\mathcal{L}_{\text{albedo}}$ is the L1 loss on the albedo
values $\{ \rho_k \}$ to encourage realistic albedo values.  Finally,
$\mathcal{L}_{\text{albedo}}$ and $\mathcal{L}_{\text{neg\_color}}$ are the
squared losses on negative diffuse color values and albedo values, respectively.  These
two loss terms penalize negative diffuse color values and albedo values,  as
negative values for these parameters are physically invalid.

The weights for the regularization losses are set as follows:
$\lambda_{\text{offset}}$ is set to 0.05, $\lambda_{\text{mask}}$,
$\lambda_{\text{normal\_orient}}$, and $\lambda_{\text{alpha\_sparsity}}$ are set to 0.1,
$\lambda_{\text{bound}}$, $\lambda_{\text{albedo}}$,
and $\lambda_{\text{neg\_color}}$ are set to 0.01. $\lambda_{\text{normal}}$ is
linearly annealed from 1.0 to 0 over the first 20k training steps.  The final
loss is defined as:
\begin{equation}
    \mathcal{L} = \mathcal{L}_{\text{rec}} + \mathcal{L}_{\text{reg}} \nonumber
\end{equation}
%

\end{document}